\documentclass[a4paper]{cas-sc}
\usepackage[numbers]{natbib}
\usepackage{graphicx}
\usepackage{booktabs}
\usepackage[ruled, vlined]{algorithm2e}
\usepackage{array}
\usepackage{times}
\usepackage{url}
\usepackage{caption}
\usepackage{multirow} 
\usepackage{amsthm, amsmath, amssymb}
\usepackage{microtype}
\usepackage{appendix}
\def\tsc#1{\csdef{#1}{\textsc{\lowercase{#1}}\xspace}}
\tsc{WGM}
\tsc{QE}

\ExplSyntaxOn
\exp_args:NNno \exp_args:Nno \use:n { \cs_gset:Npn \__make_fig_caption:nn #1#2 }
  {
    \exp_after:wN \use_ii_i:nn \exp_after:wN
      { \__make_fig_caption:nn {#1} {#2} }
      { \dim_set:Nn \l_fig_width_dim \linewidth }
  }
\exp_args:NNno \exp_args:Nno \use:n { \cs_gset:Npn \__make_tbl_caption:nn #1#2 }
  {
    \exp_after:wN \use_ii_i:nn \exp_after:wN
      { \__make_tbl_caption:nn {#1} {#2} }
      { \dim_set:Nn \l_tbl_width_dim \linewidth }
  }
\ExplSyntaxOff

\begin{document}
\let\WriteBookmarks\relax
\def\floatpagepagefraction{1}
\def\textpagefraction{. 001}

\title [mode = title]{USAD: End-to-End Human Activity Recognition via Diffusion Model with Spatiotemporal Attention} 
\shorttitle{USAD}
\shortauthors{Xiao el al.}
\author{Hang Xiao\#}
\author{Ying Yu\#}
\author{Jiarui Li}
\author{Zhifan Yang}
\author{Haotian Tang}
\author{Hanyu Liu*}
\author{Chao Li*}

\affiliation{organization={Northeastern University},
            addressline={No. 3-11, Wenhua Road, Heping District}, 
            city={Shenyang},
            postcode={110819}, 
            state={Liaoning},
            country={China}}

\begin{highlights}

\item \textbf{Addressing Key HAR Challenges with USAD:} Tackle prominent issues in Human Activity Recognition (HAR), including scarce annotations for rare activities, insufficient extraction of high-level features, and suboptimal performance in lightweight deployment scenarios. Propose a comprehensive optimization approach centered on multi-attention interaction mechanisms, aiming to mitigate these challenges.

\item  \textbf{Unsupervised Diffusion-Based Data Augmentation:} Leverage an unsupervised, statistically-guided diffusion model for data augmentation. Alleviate the scarcity of labeled data and severe class imbalance, generating high-quality synthetic samples to enrich training datasets.

\item  \textbf{Multi-Branch Spatiotemporal Interaction Network:} Design the USAD network with parallel residual branches (equipped with diverse convolutional kernels) to capture multi-scale features of sequential data. Integrate spatiotemporal attention and cross-branch feature fusion units, enhancing the overall feature representation capability for complex activity patterns.

\item  \textbf{Adaptive Multi-Loss Fusion Strategy:} Integrate an adaptive multi-loss fusion strategy. Dynamically adjust loss weights to optimize the model holistically, improving training efficacy in class-imbalanced scenarios and boosting recognition accuracy for both common and rare activities.

\end{highlights}

\begin{abstract}
The primary objective of human activity recognition (HAR) is to infer ongoing human actions from sensor data, a task that finds broad applications in health monitoring, safety protection, and sports analysis. Despite proliferating research, HAR still faces key challenges, including the scarcity of labeled samples for rare activities, insufficient extraction of high-level features, and suboptimal model performance on lightweight devices. To address these issues, this paper proposes a comprehensive optimization approach centered on multi-attention interaction mechanisms. First, an unsupervised, statistics-guided diffusion model is employed to perform data augmentation, thereby alleviating the problems of labeled data scarcity and severe class imbalance. Second, a multi-branch spatio-temporal interaction network is designed, which captures multi-scale features of sequential data through parallel residual branches with 3$\times$3, 5$\times$5, and 7$\times$7 convolutional kernels. Simultaneously, temporal attention mechanisms are incorporated to identify critical time points, while spatial attention enhances inter-sensor interactions. A cross-branch feature fusion unit is further introduced to improve the overall feature representation capability. Finally, an adaptive multi-loss function fusion strategy is integrated, allowing for dynamic adjustment of loss weights and overall model optimization. Experimental results on three public datasets—WISDM, PAMAP2, and OPPORTUNITY—demonstrate that the proposed unsupervised data augmentation spatio-temporal attention diffusion network (USAD) achieves accuracies of 98.84\%, 93.81\%, and 80.92\% respectively, significantly outperforming existing approaches. Furthermore, practical deployment on embedded devices verifies the efficiency and feasibility of the proposed method.

\end{abstract}


%%Proposed an end-to-end human activity recognition process. 
%%Utilized an unsupervised, guidance-driven diffusion model for data augmentation of human activity data. 
%%Developed the Spilt Abstraction Temporal Spatial-Dynamic Update network, which is based on the Split-Attention mechanism for multi-attentional interaction. 
%%Designed a method for real-time updating of multi-loss function weights according to batch performance. 
%\item We leverage unsupervised statistical feature-guided diffusion models for highly adaptive data augmentation
%
%\item We introduce a novel network architecture, the multi-branch spatio-temporal interaction network (MSTI), which uses multi-branch features at different levels to effectively conduct spatio-temporal interactions to enhance the ability to mine high-level latent features.
%
%\item We adopt a multi-loss function fusion strategy in the training phase to dynamically adjust the fusion weights between batches to optimize the training results.
%
%\item We built a comprehensive and detailed set of wearable deployment simulation experiments using Raspberry Pi devices qualified for today's wearable capabilities.

% Keywords
\maketitle

\section{Introduction}

Human Activity Recognition (HAR) is an emerging interdisciplinary field that aims to achieve accurate classification and dynamic tracking of human behaviors across diverse scenarios by collecting motion data and leveraging intelligent algorithms for analysis. With the growing demand for human-computer interaction and the increasing adoption of Active Assisted Living (AAL) concepts, HAR has continued to receive widespread attention.

Current HAR systems can be broadly categorized into vision-based and sensor-based approaches according to the sensing modality employed. Vision-based HAR systems recognize behaviors by capturing images or video sequences through cameras. While these systems offer the advantage of intuitive observation, they also face significant challenges, including concerns over user privacy, sensitivity to ambient lighting, limitations due to image resolution, and the high computational complexity of video processing algorithms. In contrast, sensor-based HAR systems, which utilize devices such as accelerometers and magnetometers, have gradually become the mainstream choice in both academia and industry, owing to their advantages in all-day monitoring and low power consumption.

Traditional HAR methods primarily adopt supervised learning paradigms, including classical machine learning (ML) techniques such as k-Nearest Neighbors (KNN), Decision Trees (DT), and Support Vector Machines (SVM). These algorithms are favored for their low computational complexity and broad applicability. However, their feature extraction procedures are highly dependent on expert domain knowledge—rendering the process time-consuming and labor-intensive—and they exhibit substantial limitations in adapting to small datasets and capturing complex temporal dependencies.

With advances in deep learning, Convolutional Neural Networks (CNNs), which excel at automatic extraction and hierarchical representation of spatial features, have markedly enhanced the recognition of complex human activities. Meanwhile, Recurrent Neural Networks (RNNs) demonstrate strong capabilities in temporal modeling, significantly improving the understanding of dynamic behavioral sequences. These deep learning models can automatically derive optimal discriminative feature representations from raw sensor data, thus significantly boosting recognition performance. Despite these advantages, deep learning approaches typically require large amounts of high-quality labeled data for effective training—one of the main bottlenecks in the HAR domain. 

The diversity and complexity of human activities dictate that the same activity may exhibit substantially different feature patterns in different environments. To capture such subtle distinctions and enhance recognition accuracy, contemporary HAR research extensively employs multi-source and multi-modal sensor (MMS) data. The MMS scheme enables the acquisition of richer and more complementary information regarding activities, but simultaneously presents new challenges for data fusion and feature extraction. Conventional one-dimensional convolutional networks are limited by a restricted receptive field and thus struggle to comprehensively extract spatio-temporal features from MMS data.

Compared with traditional machine learning approaches, deep learning can automatically extract high-level feature representations from sensor signals, thereby enhancing the model's ability to characterize complex motion patterns. Nevertheless, sensor-based HAR relying on deep representations still faces a series of challenges, including the scarcity of labeled data, class imbalance, and the difficulty of modeling cross-modal spatio-temporal correlations. This paper, therefore, makes three main contributions:

\textbf{A Diffusion Model-Based Approach for Imbalanced Data Processing:} To address the challenges of limited labeled data and severe class imbalance, we first employ an unsupervised diffusion model guided by statistical features to achieve high-quality data augmentation. This model is capable of maintaining the original statistical properties while generating diverse synthetic samples, effectively expanding the training dataset for rare categories.

\textbf{Multi-Branch Spatiotemporal Interaction Network (USAD):} We propose a novel network architecture that incorporates multi-branch residual structures with deep spatio-temporal interaction mechanisms. Through multi-branch design, the architecture processes and encodes spatial and temporal features from different scales and perspectives, thereby enhancing the diversity and expressive power of features.

\textbf{Comprehensive Validation and Practical Deployment:}  We conduct a thorough evaluation on three public HAR benchmark datasets, demonstrating that the proposed approach consistently outperforms state-of-the-art methods across various metrics. Detailed ablation studies quantify the contributions of each component and further corroborate the effectiveness of our method.

The remainder of this paper is organized as follows: Section 2 systematically reviews related research in HAR, including deep feature extraction methods, data augmentation techniques, and the application of attention mechanisms. Section 3 details the proposed multi-branch spatio-temporal interaction network, data augmentation strategy, and composite loss function. Section 4 describes the experimental settings, including datasets, evaluation metrics, and comparative methods. Section 5 presents and discusses the experimental results, summarizes the research findings, and outlines directions for future work.

\section{Related work}

\subsection{Deep Feature Extraction}

High-quality feature extraction is fundamental to accurate complex human activity classification and recognition. Recent research on neural network architecture design and multi-source HAR feature integration has advanced significantly, yet conventional ResNet architectures still suffer from weak inter-channel correlations, parameter redundancy, and insufficient feature reutilization.. To address these, Zhang et al.. \cite{zhang2023attention} proposed ResNeSt, introducing a Split-Attention module and restructured feature aggregation to enhance channel dependency modeling and feature representation. Mekruksavanich et al. \cite{benmessaoud2025sensor} adopted ResNeXt, integrating inertial and stretch sensor data for efficient activity classification. Other studies enrich feature extraction methods: Y. Wang et al. \cite{wang2023novel} proposed DMEFAM, integrating temporal-channel-spatial features via Bi-GRU, self-attention, CBAM, and ResNet-18, improving accuracy and DAAD generalization. Zhang J. et al. \cite{zhang2024diverse} used contrastive learning to extract high-precision activity-style features from multi-source data, preserving data characteristics and class semantics for enhanced cross-subject generalization. Liu X. et al. \cite{zhao2025cir} encoded time series into RGB images via MTF/RP/GAF, designing CIR-DFENet with multi-scale convolution, global attention, and 1D CNN+LSTM for effective multi-source feature fusion.

\subsection{Data Augmentation}

A variety of data augmentation methods have been proposed to address data scarcity and model overfitting. Early work by J. Wang \cite{wang2018sensorygans} introduced SensoryGANs, leveraging generative adversarial networks (GANs) to generate synthetic sensor data for improved activity recognition accuracy, alongside visualization methods for evaluating synthetic data quality. However, SensoryGANs faced challenges including unstable training dynamics and the requirement for activity-specific models, which significantly reduced computational efficiency. Li, X \cite{li2020activitygan} proposed ActivityGAN to generate multi-activity data within a unified framework, effectively expanding datasets and enhancing model training efficacy and classification accuracy. Nevertheless, it failed to yield substantial improvements in recognition performance and exhibited notable limitations in cross-activity generalization capabilities.

To overcome these limitations of traditional approaches, researchers have turned to denoising diffusion probabilistic models (DDPM) \cite{ho2020denoising}. S. Shao \cite{shao2023study} designed a diffusion modeling framework tailored to sensor data characteristics, reconstructing the UNet architecture to handle low-channel and variable-length sequences, simplifying sampling layers to avoid over-feature extraction, and embedding class labels for conditional generation. Experimental results on HASC and PAMAP2 datasets demonstrated that the quality of generated data surpassed traditional transformation methods and GAN-based approaches, with a 1:4 real-synthetic data blending ratio further optimizing model performance. Current diffusion-based methods predominantly focus on unimodal data (e.g., inertial sensors), while research on collaborative augmentation for multimodal data (e.g., IMU sensors) in human activity recognition (HAR) tasks remains insufficient. Although Zhou, Y \cite{zhou2024autoaughar} adopted gradient optimization to determine augmentation strategy probability distributions, enabling automatic and efficient policy search—operating in two stages, accommodating multimodal properties, and demonstrating model-agnostic scalability—it exhibited limited performance gains in adjacent domains (e.g., EEG classification), highlighting that modality specificity inherently constrains the universality of augmentation strategies

\subsection{Attention Mechanisms}

 In recent years, attention-based deep learning models have gained traction. Abedini, A. \cite{mekruksavanich2023hybrid} proposed Att-CNN-BLSTM, using CNN and BLSTM to extract spatial-temporal features with an attention layer, achieving 90.54\% accuracy and 86.88\% F1-score, outperforming traditional models, but with high computational complexity and limited interpretability. Ye, X. et al. \cite{ye2024deep} developed DGDA, combining generative models with temporal attention to exploit time series dependencies, improving distribution alignment for HAR with excellent cross-user performance, yet limited in fusing heterogeneous sensors and resource-heavy. Zhao, J. et al. \cite{Zhao2023CNNAttBiLSTM} integrated self-attention into CNN-BiLSTM for DDoS detection, achieving 95.67\% accuracy via 1D CNN-BiLSTM feature extraction and attention emphasis, but with increased complexity needing efficiency optimization. Zhang, X. \cite{akter2023human} designed AM-DLFC with 2D CNN and CBAM, lightweight and generalizable, but prone to misclassifying specific actions and ignoring real-world interferences.

\section{Methodology}

\begin{figure}[!ht]
    \centering
    \includegraphics[width=0.7\textwidth]{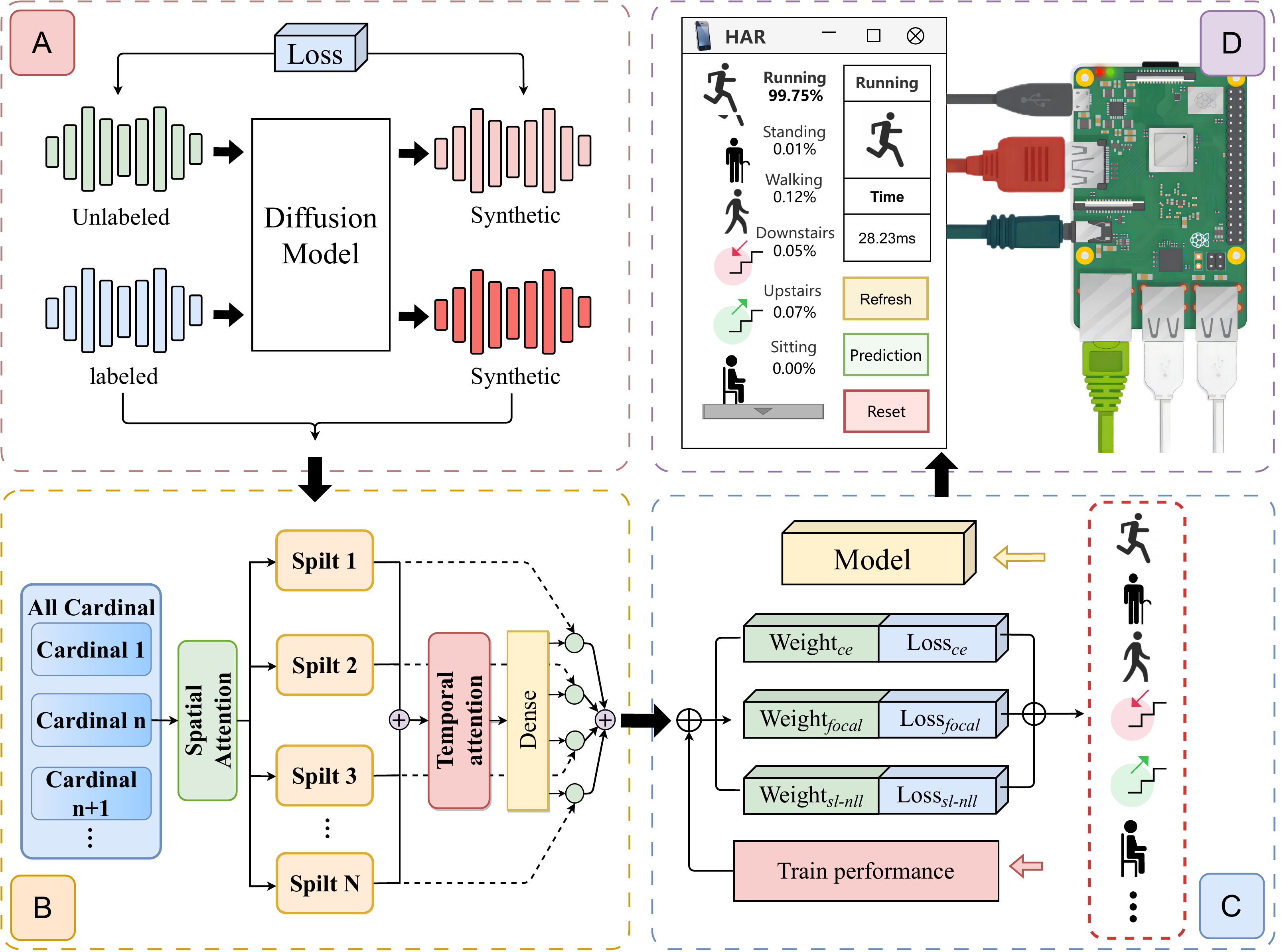}
    \caption{Framework of our USAD method, including (A) Data synthesis, (B) Multi-branch network architecture, (C) Adaptive composite loss function, (D) Embedded device deployment.}
    \label{Figure_1}
\end{figure}

This section introduces the proposed HAR framework. As shown in Figure \ref{Figure_1}, it consists of four parts: Data augmentation based on unsupervised diffusion model, Multi-brach network architecture of spatiotemporal attention interaction, Adaptive composite loss function and Embedded device deployment.

\subsection{Sensor Data Augmentation Framework Based on Conditional Diffusion Models}

We propose an innovative sensor data augmentation framework that substantially enhances human activity recognition (HAR) systems by generating high-fidelity synthetic sequences. As illustrated in Figure~\ref{Figure_1}, the proposed framework consists of three core stages: (1) conditional diffusion model pretraining; (2) classifier pretraining on synthetic data; and (3) classifier fine-tuning on real-world data. In this section, we present the mathematical formulation and architectural design of each stage in detail.

\subsubsection{Statistical Feature Extraction and Label Conditioning}

Given a sequence of sensor time-series data $x_0 \in \mathbb{R}^L$, where $L$ denotes the sequence length, we extract multi-granularity statistical features and establish a conditioning mechanism with the associated activity label $y$.

Global statistical features are computed as follows:
\begin{equation}
\mu = \frac{1}{L} \sum_{i=1}^{L} x_{0,i}, \quad
\sigma = \sqrt{ \frac{1}{L} \sum_{i=1}^{L} (x_{0,i} - \mu)^2 }, \quad
\gamma = \frac{1}{L} \sum_{i=1}^{L} \left( \frac{x_{0,i} - \mu}{\sigma} \right)^3
\end{equation}

Local point-wise features are extracted using Z-score normalization:
\begin{equation}
z_i = \frac{x_{0,i} - \mu}{\sigma}, \quad \forall i \in \{1, \dots, L\}
\end{equation}

We construct a feature-conditioned vector by replicating the global features:
\begin{equation}
f = 
\begin{bmatrix}
\underbrace{\mu, \dots, \mu}_{L},\ 
\underbrace{\sigma, \dots, \sigma}_{L},\ 
\underbrace{\gamma, \dots, \gamma}_{L},\ 
z_1, \dots, z_L
\end{bmatrix}^\top \in \mathbb{R}^{4L}
\end{equation}

To associate features with class labels, we compute label-specific prototype statistics:
\begin{equation}
\mu_y = \mathbb{E}[f \mid y] = \frac{1}{N_y} \sum_{j: y^{(j)} = y} f^{(j)}
\end{equation}
where $N_y$ is the number of training samples for class $y$. The vector $\mu_y$ is used to guide conditional generation during the synthesis process.

\subsubsection{Conditional Diffusion Model Pretraining}

\textbf{Forward Diffusion Process:}
We adopt a standard DDPM-based forward process:
\begin{equation}
x_t = \sqrt{\bar{\alpha}_t} x_0 + \sqrt{1 - \bar{\alpha}_t} \, \epsilon, \quad
\epsilon \sim \mathcal{N}(0, I), \quad
t \sim \mathcal{U}\{1, T\}
\end{equation}
where $\bar{\alpha}_t = \prod_{s=1}^{t} (1 - \beta_s)$. The noise schedule is defined by a cosine-based formulation:
\begin{align}
\bar{\alpha}_t &= \frac{\cos\left( \frac{\pi}{2} \cdot \frac{t/T + s}{1 + s} \right)}{\cos\left( \frac{\pi}{2} \cdot \frac{s}{1 + s} \right)} \\
\beta_t &= \text{clip}\left(1 - \frac{\bar{\alpha}_t}{\bar{\alpha}_{t-1}},\ 0,\ 0.999\right)
\end{align}
with an offset $s = 0.008$ to ensure $\bar{\alpha}_0 \approx 1$.

\textbf{Label-Conditioned Denoising via Adaptive Normalization:}
The reverse denoising process is parameterized by a conditional generator $G_\theta$:
\begin{equation}
\hat{\epsilon}_\theta = G_\theta(x_t,\ t,\ f,\ y)
\end{equation}

To integrate label conditioning, we use adaptive group normalization (AdaGN) in each residual block:
\begin{equation}
\text{AdaGN}(h,\ t,\ y) = \gamma_y(t) \odot \frac{h - \mu_h}{\sigma_h} + \beta_y(t)
\end{equation}

The parameters $\gamma_y(t), \beta_y(t)$ are generated by a multilayer perceptron (MLP) from label and time embeddings:
\begin{equation}
\begin{bmatrix}
\gamma_y(t) \\
\beta_y(t)
\end{bmatrix}
= \text{MLP} \left( \text{concat}[e_y,\ \psi(t)] \right)
\end{equation}
where $\psi(t) \in \mathbb{R}^{128}$ is a sinusoidal time embedding and $e_y$ is a learnable label embedding.

\textbf{Training Objective:}
The model is trained by minimizing the weighted mean squared error between predicted and true noise:
\begin{equation}
\mathcal{L}_{\text{rec}}(\theta) = \mathbb{E}_{x_0,\ t,\ \epsilon,\ y} \left[
w_t \left\| \epsilon - G_\theta(x_t,\ t,\ f,\ y) \right\|_2^2
\right]
\end{equation}
with the importance weight:
\begin{equation}
w_t = \sqrt{ \frac{1 - \bar{\alpha}_t}{\bar{\alpha}_t (1 - \bar{\alpha}_{t-1})} }
\end{equation}

\subsubsection{Classifier Pretraining with Synthetic Data}

\textbf{Conditional Data Sampling:}
After fixing the pretrained generator $G_{\theta^*}$, we sample synthetic sequences with class-balanced labels:
\begin{align}
x_{\text{syn}}^{(k)} &= G_{\theta^*}(\omega^{(k)},\ \mu_{y^{(k)}},\ y^{(k)}), \\
\omega^{(k)} &\sim \mathcal{N}(0, I), \quad
y^{(k)} \sim \mathcal{U}\{1,\ n_c\}
\end{align}

We obtain the synthetic dataset:
\begin{equation}
\mathcal{D}_{\text{syn}} = \left\{ \left(x_{\text{syn}}^{(k)},\ f_{\text{syn}}^{(k)},\ y^{(k)}\right) \right\}_{k=1}^M
\end{equation}
with $f_{\text{syn}}$ computed from the generated sequence.

\textbf{Multimodal Classifier Architecture:}
The HAR classifier $C_\phi$ processes 5-channel temporal input formed by concatenating the raw signal $x$ and feature vector $f$:
\begin{equation}
C_\phi([x; f]): \mathbb{R}^{5 \times L} \rightarrow \Delta^{n_c - 1}
\end{equation}

The network consists of three 1D convolutional blocks (kernel size = 5, stride = 2, channels = 64, GeLU activation), a global average pooling layer, two fully connected layers (128 units, Dropout=0.3), and a Softmax output layer.

\textbf{Pretraining Objective:}
We minimize the cross-entropy loss on the synthetic dataset:
\begin{equation}
\mathcal{L}_{\text{syn}}(\phi) = -\frac{1}{M} \sum_{k=1}^M \sum_{c=1}^{n_c} y_c^{(k)} \log C_\phi\left( [x_{\text{syn}}^{(k)};\ f_{\text{syn}}^{(k)}] \right)_c
\end{equation}

The pretrained parameters $\phi_{\text{init}}$ serve as initialization for fine-tuning.

\subsubsection{Classifier Fine-Tuning on Real Data}

To bridge the domain gap between synthetic and real data, we fine-tune $C_\phi$ on the real dataset $\mathcal{D}_{\text{real}} = \{(x_0^{(j)}, f^{(j)}, y^{(j)})\}_{j=1}^N$:
\begin{equation}
\mathcal{L}_{\text{real}}(\phi) = -\frac{1}{N} \sum_{j=1}^N \sum_{c=1}^{n_c} y_c^{(j)} \log C_\phi\left( [x_0^{(j)};\ f^{(j)}] \right)_c
\end{equation}

Theoretically, if the synthetic distribution $p_{\text{syn}}$ approximates the real distribution $p_{\text{real}}$, the risk difference is bounded by:
\begin{equation}
\left| \mathbb{E}_{x \sim p_{\text{syn}}} \mathcal{L}(C_\phi(x)) - \mathbb{E}_{x \sim p_{\text{real}}} \mathcal{L}(C_\phi(x)) \right| \leq \mathcal{O}\left( W(p_{\text{syn}},\ p_{\text{real}}) \right)
\end{equation}
where $W(\cdot,\cdot)$ denotes the Wasserstein distance.

\subsection{Multi-Branch Network Architecture with Spatiotemporal Attention Interaction}

\begin{figure}[!ht]
    \centering
    \includegraphics[width=0.7\linewidth]{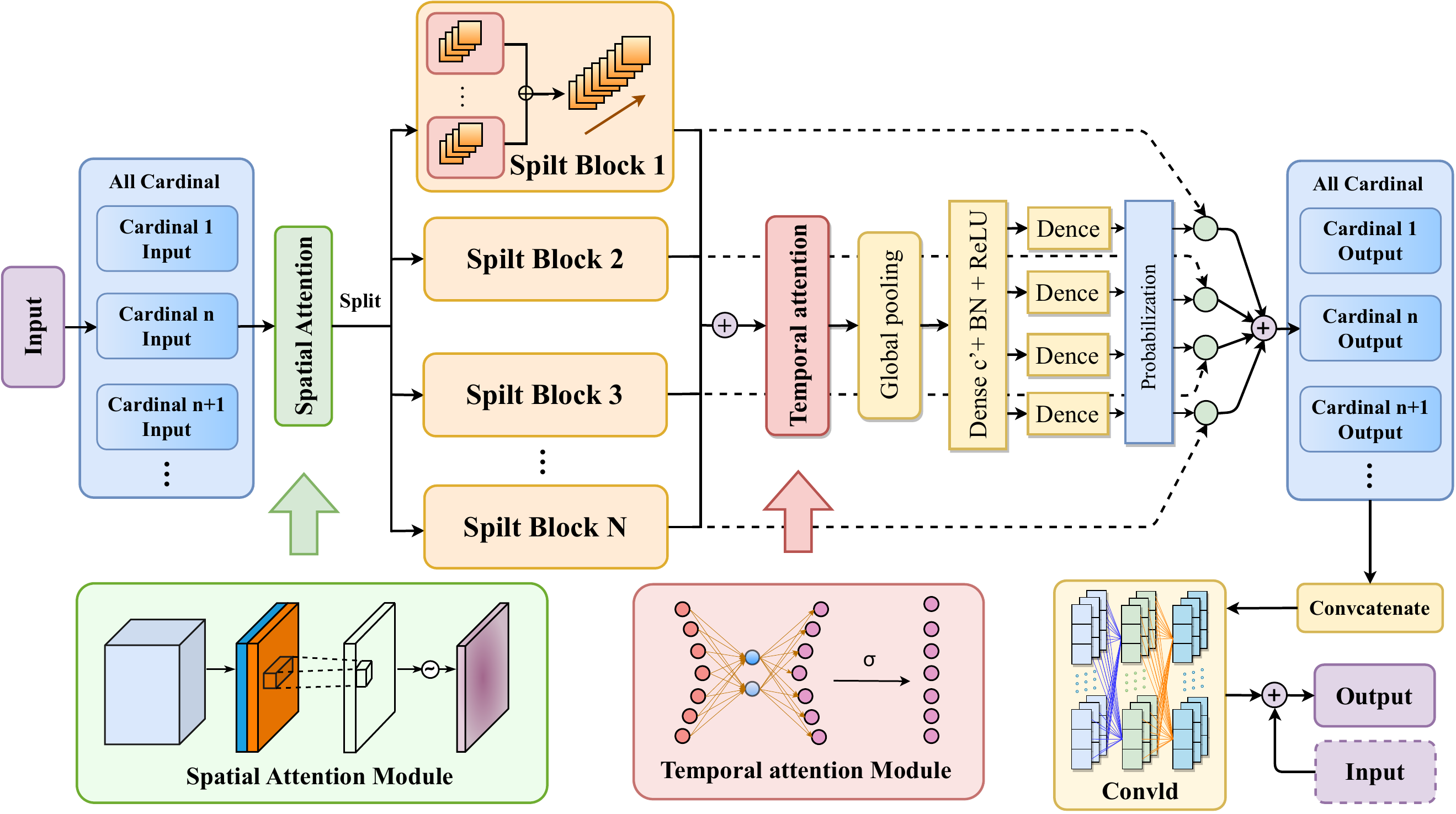}
    \caption{Detailed structure of the proposed USAD model}
    \label{Figure_2}
\end{figure}

Branch attention models can enhance accuracy without significantly increasing computational cost. Based on this, the present study proposes the USAD model, which achieves improved computational efficiency through optimization of the number of nonlinear branches, the number of intermediate channels, and network structure. Building upon this foundation, temporal and spatial attention mechanisms are further incorporated for more precise capture of data spatiotemporal characteristics. Consequently, USAD enhances feature extraction capability while improving computational efficiency. Figure \ref{Figure_2} illustrates the architecture of the USAD model.

USAD employs a multi-branch convolutional architecture, focusing on the spatial and temporal properties at different time segments. By decomposing and reorganizing features, it optimizes the specialization and diversity of feature extraction as well as the expressiveness of the network. The generated data and features presented previously are directly inserted. The features are partitioned into multiple groups, where the number of feature groups is determined by the hyperparameter $K$,  and the hyperparameter $R$ specifies the number of splits within each cardinal group. Thus, the total number of feature groups is $G = KR$. Each group undergoes a series of transformations, denoted as $\{F_1, F_2, ..., F_n\}$, with the intermediate representation of the $i$-th group given by $U_i = F_i(X)$, for $i \in \{1, 2, ..., G\}$, where $X$ represents the input features.

\textbf{Cardinality:}
The cardinality module, inspired by ResNeSt, integrates group features at the element level by summing elements across groups, thereby merging multi-scale information and further enhancing global contextual information integration. This improves the expressive power of the features. Specifically, the feature representation of the $k$-th cardinal group is $\hat{U}^k \in \mathbb{R}^{S \times C / K}$ for $k \in 1,2, \ldots K$, generated by element-wise summation over all splits $U_j$ in the group:
\begin{equation}
{U}^k=\sum_{j=R(k-1)+1}^{R k} U_j, 
\end{equation}
where $S$ and $C$ represent the spatial size and channel dimension of the feature map, respectively.

To further capture global contextual information, global average pooling is performed across spatial dimensions to obtain descriptive statistics $D_c^k \in \mathbb{R}^{K}$. The statistic for the $c$-th channel is computed as $D_c^k=\frac{1}{S} \sum_{j=1}^S \hat{U}_c^k(i, j)$. This design not only ensures the richness of feature representations but also provides vital support for subsequent attention mechanisms.

\textbf{Radix:}
The radix split module employs channel-wise soft attention to effectuate weighted fusion of features, further improving feature discrimination and representational power. Specifically, the final output for each channel $V^k \in \mathbb{R}^{S \times C / K}$ is obtained by weighted summation of the corresponding split features: $V_c^k=\sum_{i=1}^Ra_i^k(c) U_{R(k-1)+i}$, where $a_i^k(c)$ is a soft assignment weight controlling the contribution of each split feature to the final output, calculated as:
\begin{equation}
a_i^k(c)= \begin{cases}\frac{\exp \left(G_i^c\left(s^k\right)\right)}{\sum_{j=1}^R\left(\exp \left(G_j^c\left(s^k\right)\right)\right.} & \text { if } R>1, \\ \frac{1}{1+\exp \left(-G_i^c\left(s^k\right)\right)} & \text { if } R=1,\end{cases}
\end{equation}
The soft attention mechanism adaptively adjusts the feature weights, emphasizing salient features while suppressing irrelevant ones, thereby enhancing the fusion effect and improving the adaptability to diverse features. Here, $V^k$ denotes the version of $U^k$ after soft attention weighting and fusion.

\textbf{Feature Concatenation:}
The feature concatenation module concatenates features of multiple basis groups along the channel dimension. Residual connections are applied to preserve the information flow of the input features, simultaneously enriching the diversity of feature fusion and maintaining the integrity of key information. This effectively enhances the accuracy and robustness of feature representation. The concatenation operation is expressed as: $V = \text{Concat}\{V_1, V_2, ..., V_K\}$, where $\{V_1, V_2, ..., V_K\}$ denote the representations of each basis group. 

\textbf{Method Flow:}
This method incrementally optimizes feature extraction and fusion via a multi-branch structure combined with attention mechanisms. The input feature map is divided into $KR$ groups, each tagged by a $K$ and an $R$ index, with groups having the same cardinality index arranged adjacently along the channel dimension. On the basis of channel-level soft attention for each basis group, a spatial attention mechanism is further introduced to enhance the fine-grained feature representation ability.

First, global temporal features are extracted by performing average and max pooling along the spatial direction and concatenating the results to form the descriptor $F=[F_{avg}^{s};F_{max}^{s}]$. A convolutional layer generates the spatial attention map $M_s(F)\in \mathbb{R}^{S\times C}$ on the concatenated feature, encoding spatial locations of interest in each map.

Next, a cross-split summation is performed to fuse feature maps with identical $K$ but different $R$ indices. Subsequently, a temporal attention mechanism is applied to reduce dependencies among groups with different  $R$ indices.

Temporal attention compresses the spatial dimension $S$ via a global pooling layer, yielding the channel descriptor $Z_c \in \mathbb{R}^C$, where the statistic for the $c$-th channel is:
\begin{equation}
Z_c=\frac{1}{S} \sum_{i=1}^S u_c(i, j)
\end{equation}
Here,$u_c(i, j)$ denotes the value of the $c$-th channel at spatial location $(i, j)$. The channel-wise statistics extracted by global pooling are used for subsequent attention allocation. The pooling layer maintains feature separation along the channel dimension, i.e., global pooling is carried out independently for each basis group. The pooled results are processed by two consecutive fully connected layers, each with a number of groups equal to the cardinality, to predict the attention distribution of each split module.

For further optimization, previous $1\times 1$ convolutions are unified into a single grouped convolutional layer, while a $3\times 1$ convolution is used for feature fusion, increasing the grouping number to $K\times R$. Ultimately, the decentralized attention blocks are modularized together with standard CNN modules, with residual connections and concatenation further strengthening feature expression and modeling, thereby achieving an efficient feature extraction and fusion pathway.

\subsection{Adaptive Composite Loss Function}

The class imbalance problem is highly prevalent in sensor-based human activity recognition tasks and severely impedes both the training and generalization capabilities of classifiers. To address this challenge, this paper proposes an adaptive composite loss function that integrates cross-entropy loss, focal loss, and label smoothing regularization, thereby enhancing the model's adaptability and robustness with respect to imbalanced data distributions.

Specifically, the cross-entropy loss serves as the fundamental component for multi-class classification tasks, directly optimizing the model’s decision boundaries by minimizing the divergence between the predicted and true class distributions. Its mathematical formulation is:
\begin{equation}
Loss_{ce}=-\sum_{i=1}^p\left(x_{i}\right)\mathrm{log}q\left(x_{i}\right)
\end{equation}
where this term measures the discrepancy between the model’s outputs and the true distribution, guiding the model to learn accurate class separation.

To further improve the identification of minority classes, the focal loss is introduced. By assigning greater weights to hard-to-classify samples, focal loss dynamically adjusts the contribution of easy and difficult samples in the total loss. The focal loss is defined as:
\begin{equation}
Loss_{fl}=-\alpha_{t}(1-p_{t})^{\gamma}\log{(p_{t})}
\end{equation}
where $\alpha_{t}$ denotes the sample weight factor and $\gamma$ is the focusing parameter. This loss function effectively increases the model’s attention to minority class samples and mitigates the performance bottleneck caused by class imbalance.

Additionally, to prevent the model from becoming overconfident in any single class during training and to improve generalization, label smoothing regularization is employed. By smoothing the label distribution and introducing a minor perturbation, label smoothing reduces the risk of overfitting to the training data. Its formulation is as follows:
\begin{equation}
Loss_{sl-nll}=-\sum_{k=1}^K\log p(k)\left((1-\epsilon)\delta_{k, y}+\frac\epsilon K\right)
\end{equation}
where $\epsilon$ is the smoothing parameter and $K$ is the total number of classes. This term encourages a more balanced model output distribution, thereby enhancing generalization to unseen data.

At the conclusion of each training batch, these three loss functions are combined via weighted summation to form the final composite loss function, leveraging their respective strengths for robust adaptation to complex data distributions. The combined loss is expressed as:
\begin{equation}
Loss_{total}=\omega_{0}Loss_{sl-nll}+\omega_{1}Loss_{fl}+\omega_{2}Loss_{ce}
\end{equation}
where $\omega _0$, $\omega_1$, $\omega_2$ represent the weights assigned to the respective loss components. This weighting mechanism allows flexible balancing of the contributions from different losses, thereby enhancing overall model performance.

Given the challenge of manually tuning weights among multiple loss terms, an adaptive weighting algorithm is further proposed in this paper. Inspired by the feedback principle in automatic control theory, the algorithm dynamically adjusts the loss weights in each training epoch according to model performance. Specifically, if the current epoch's model accuracy exceeds that of the previous epoch, the higher weight is slightly reduced and the lower weight is slightly increased, and vice versa. The update rules for weights are as follows:
\begin{equation}
\begin{aligned}
\omega_0&=0. 5*(1-\omega_1)\\
\omega_1&=2-\tau-\frac1{acc+1e-8}\\
\omega_2&=0. 5*(1-\omega_1)\\
\end{aligned}
\end{equation}
where $\tau$ is the weight-tuning parameter and $acc$ denotes the current model accuracy. This scheme enables automatic adjustment of each loss term’s impact based on model performance, ensuring that the loss function remains dynamically balanced throughout training.

\subsection{Embedded Device Deployment}  

In recent years, many studies have evaluated models exclusively on high-performance devices; however, when deploying HAR models in practical scenarios, various resource constraints inherent to target devices must often be addressed. To this end, this work designs and implements an embedded deployment system aligned with the current state of wearable technology, aiming to assess the feasibility of various approaches in real-world applications. The proposed deployment solution draws on the work of Zhang et al., with experiments conducted on a Raspberry Pi 5 platform to compare the inference latency of the proposed modules against baseline models. Furthermore, we construct a more comprehensive and detailed deployment evaluation framework, the specific implementation details of which are elaborated upon in the "Practical Deployment" section of the Experiments.

\section{Results \& Discussion}

In this section, we present a systematic and in-depth empirical analysis of the proposed USAD method. Specifically, we comprehensively evaluate the method from multiple perspectives, including the effectiveness of the model architecture, its adaptability to practical challenges, the impact of key mechanism parameters (Appendix C), and overall performance comparison.

First, ablation studies are conducted to quantify the independent contribution of each core module, thereby validating the rationality of the model design. Subsequently, we systematically analyze the performance of USAD in addressing the common issue of class imbalance in real-world HAR tasks. Furthermore, through experiments on the hyperparameters of the adaptive composite loss function, we verify the effect of the dynamic loss adjustment strategy on both overall model performance and the recognition ability for minority classes. Finally, large-scale comparative experiments are conducted between USAD and current mainstream methods on the WISDM dataset and the PAMAP2 dataset (with 100\% and 50\% data volumes), objectively demonstrating the superior performance of USAD across multiple public datasets.

\subsection{Ablation Study}

To comprehensively validate the effectiveness of each module in the proposed method across different types of datasets, as shown in Table \ref{Table_1}, we conducted ablation experiments on the WISDM, PAMAP2, and OPPORTUNITY datasets. These three datasets exhibit increasing complexity in terms of individual distribution differences and class imbalance, thus posing greater challenges for evaluating the contribution of each model component. In this experiment, the USAD architecture was systematically dissected to assess the practical effect of each core component.

% 第一个表格
\begin{table}[htbp]
\centering
%\renewcommand{\arraystretch}{1.1}
%\resizebox{\textwidth}{!}{
%\fontsize{10pt}{12pt}\selectfont
\resizebox{\textwidth}{!}{
\begin{tabular}{llcccccc}
\toprule
\multicolumn{2}{l}{\textbf{Model}} & \multicolumn{6}{c}{\textbf{WISDM}} \\
\cmidrule(lr){3-8}
\textbf{Main} & \textbf{Additional} & \textbf{Acc. (\%)} & \textbf{Pre. (\%)} & \textbf{Rec. (\%)} & \textbf{F1 (\%)} & \textbf{G-mean (\%)} & \textbf{AUC (\%)} \\
\midrule
\multirow{5}{*}{\textbf{Base}} 
& Backbone Only & 97.74 & 96.87 & 97.49 & 97.15 & 98.55 & 97.74 \\
& Spatial Attention & 97.78 & 97.43 & 97.77 & 97.59 & 98.70 & 98.64 \\
& Temporal Attention & 97.87 & 97.48 & 97.64 & \textbf{98.58} & 97.54 & 98.59 \\
& Spatial and Temporal Attention & 97.96 & 97.75 & 97.90 & 97.82 & 98.77 & 98.72 \\
& ACL & 98.01 & \textbf{97.99} & 97.83 & 97.87 & 98.74 & 98.70 \\
\midrule
\multirow{3}{*}{\textbf{Fixed Weights}} 
& Compound Loss & 97.38 & 95.70 & 97.25 & 96.42 & 98.40 & 98.36 \\
& Combinations 1 & 97.92 & 96.92 & 97.92 & 97.40 & 98.78 & 98.74 \\
& Combinations 2 & 97.83 & 97.56 & 97.55 & 97.55 & 98.59 & 98.54 \\
\midrule
\textbf{USAD} 
& Data Augmentation & \textbf{98.28} & 97.94 & \textbf{98.39} & 98.16 & \textbf{99.05} & \textbf{99.01} \\
\bottomrule
\end{tabular}}
%}
\caption{Ablation Study on WISDM}
\label{tab:ablation1}
\end{table}
% 续表
\begin{table}[htbp]
\ContinuedFloat
\centering
\resizebox{\textwidth}{!}{
%\fontsize{10pt}{12pt}\selectfont
\begin{tabular}{cccccc|cccccc}
\toprule
\multicolumn{6}{c|}{\textbf{PAMAP2}} & \multicolumn{6}{c}{\textbf{OPPORTUNITY}} \\
\cmidrule(lr){1-6} \cmidrule(lr){7-12}
\textbf{Acc. (\%)} & \textbf{Pre. (\%)} & \textbf{Rec. (\%)} & \textbf{F1 (\%)} & \textbf{G-mean (\%)} & \textbf{AUC (\%)} 
& \textbf{Acc. (\%)} & \textbf{Pre. (\%)} & \textbf{Rec. (\%)} & \textbf{F1 (\%)} & \textbf{G-mean (\%)} & \textbf{AUC (\%)} \\
\midrule
90.71 & 90.89 & 90.14 & 90.27 & 94.14 & 94.19 & 78.05 & 71.83 & 72.23 & 71.42 & 82.44 & 83.08 \\
90.61 & 91.04 & 90.49 & 90.61 & 94.32 & 94.36 & 75.02 & 69.59 & 70.53 & 69.39 & 79.39 & 82.13 \\
90.52 & 90.40 & 90.16 & 90.13 & 94.14 & 94.19 & 77.39 & 68.99 & 70.59 & 69.08 & 81.49 & 82.25 \\
92.00 & 91.57 & 92.08 & 91.70 & 95.64 & 95.67 & 81.95 & 76.18 & 76.40 & 75.54 & 86.12 & 86.69 \\
93.43 & 93.59 & 93.19 & 93.10 & 96.27 & 96.29 & 83.11 & 76.37 & 75.54 & 75.44 & 86.08 & 86.77 \\
\midrule
93.24 & 93.83 & 92.55 & 93.01 & 95.93 & 95.96 & 83.28 & \textbf{78.64} & 78.62 & 78.10 & 87.82 & 88.32 \\
91.43 & 91.34 & 91.54 & 91.29 & 95.33 & 95.37 & 83.03 & 76.13 & 75.98 & 75.38 & 86.32 & 86.99 \\
89.81 & 90.38 & 90.13 & 89.93 & 94.53 & 94.60 & 83.54 & 75.93 & 76.79 & 75.76 & 86.77 & 87.38 \\
\midrule
\textbf{94.07} & \textbf{94.34} & \textbf{93.33} & \textbf{93.72} & \textbf{96.37} & \textbf{96.39} & \textbf{84.60} & 78.47 & \textbf{79.85} & \textbf{78.46} & \textbf{88.95} & \textbf{89.44} \\
\bottomrule
\end{tabular}
}
\caption{Ablation Study on PAMAP2 and OPPORTUNITY (continued)}
\label{Table_1}
\end{table}

The baseline model (Backbone Only) does not include any additional attention mechanisms and consists only of the multi-branch structure with cardinality and cardinality grouping. Subsequently, we introduced spatial attention and temporal attention modules separately, and further compared their effects with the simultaneous integration of both (i.e., the complete spatiotemporal interaction module) on model performance. "ACL" denotes the training strategy with the adaptive weighted optimization algorithm, aiming to examine the impact of dynamic weight adjustment mechanisms on model performance. "Data Augmentation" refers to the addition of data augmentation based on an unsupervised diffusion model, which constitutes the final USAD model. In addition, "Compound Loss" indicates the use of an unweighted triple-loss combination optimization strategy during model training. To more intuitively demonstrate the individual impact and contribution of the three loss functions to overall performance, we further evaluated two sets of optimal weighted combinations: in Combination 1, the weights are set to 0.3, 0.4, and 0.3, respectively; in Combination 2, the weights are set to 0.15, 0.7, and 0.15, respectively.

The baseline model already achieves high accuracy and G-mean on the WISDM dataset, demonstrating that the multi-branch structure is well-suited for relatively simple or class-balanced data. However, as the dataset complexity increases, the performance of the baseline model declines significantly. This indicates that the model’s generalization ability is limited in scenarios characterized by class imbalance and high complexity. Building upon this, we separately introduced spatial attention and temporal attention modules. Experimental results show that incorporating either spatial or temporal attention alone yields limited improvements in overall model performance, and in some metrics, even leads to slight declines. This phenomenon suggests that attention mechanisms focusing on a single dimension may result in incomplete feature capture or even introduce noise, thereby impairing the model’s ability to model complex spatio-temporal relationships. In contrast, the spatio-temporal attention interaction module enables more comprehensive feature fusion, resulting in significant improvements across all metrics on the three datasets. These results validate the importance of multi-dimensional feature interaction for complex activity recognition tasks.

With the introduction of the ACL, the model’s performance is further enhanced across all datasets, indicating that the dynamic loss weighting mechanism effectively mitigates the negative impact of class imbalance and improves the model’s robustness and generalization. In the final USAD model, the unsupervised diffusion model-based data augmentation strategy is incorporated, leading to optimal performance on all three datasets. Specifically, the final model improves accuracy by 3.36\% on the PAMAP2 dataset and by 6.55\% on the OPPO dataset. Data augmentation not only enhances the model’s ability to recognize minority classes but also further strengthens the overall feature representation.

To further analyze the impact of loss function design on model performance, we compared the unweighted triple loss combination with two sets of optimal weighted combinations. The results show that the fixed-weight combinations on model performance vary from different datasets, indicating that reasonable adaptive loss weight allocation can improve the model’s adaptability in different scenarios.

In summary, each module of USAD demonstrates positive effects across different types of datasets. The multi-branch structure provides a solid foundation for feature representation, the spatio-temporal attention interaction mechanism significantly enhances the model’s ability to recognize complex activities, and adaptive loss weighting and unsupervised data augmentation further improve the model’s generalization and robustness. Ultimately, USAD achieves optimal comprehensive performance on all three datasets, validating the effectiveness and generalizability of the proposed method for diverse human activity recognition tasks.

\subsection{Study on Addressing Class Imbalance}

To systematically validate the effectiveness of the proposed USAD method in addressing class imbalance, experiments were conducted on three public datasets: WISDM (mild imbalance), PAMAP2 (moderate imbalance), and OPPORTUNITY (severe imbalance). The focus was on evaluating the impact of class imbalance on model performance, using both quantitative metrics (Appendix B) and T-SNE visualization to compare the performance of USAD with various mainstream models. By benchmarking against classical CNN, recurrent neural networks (LSTM), basic residual networks (ResNet), attention-based models (SE-Res2Net), as well as high-performance existing models (MAG-Res2Net, ATFA\cite{FENG2024121296}, TCCSNet\cite{ESSA2023110867}), we assessed the strengths and weaknesses of each method from the perspectives of classification performance and feature space distribution.

\subsubsection{Comparative Analysis of Classification Performance}

\begin{figure}[htbp]
    \centering
    \includegraphics[width=\textwidth]{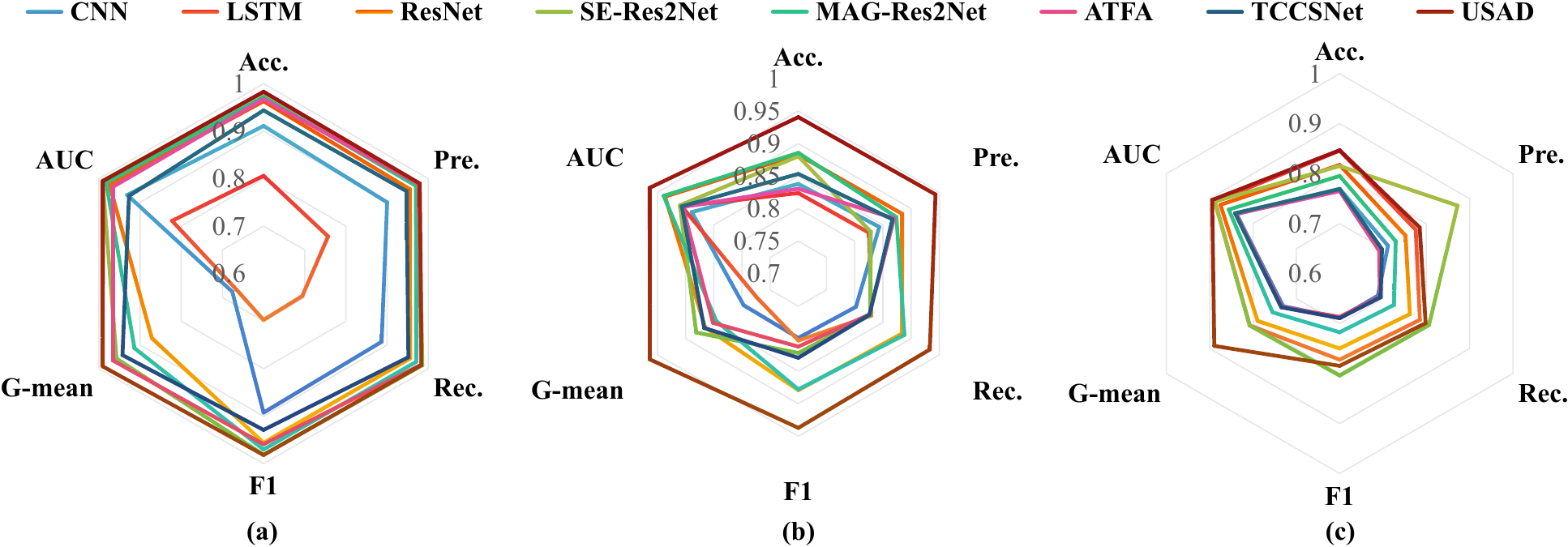}
    \caption{(a) shows the model performance comparison on the WISDM dataset, (b) on the PAMAP2 dataset, and (c) on the OPPORTUNITY dataset. The radar charts use Acc., Pre., Rec., F1, G - mean, and Area Under the Curve (AUC) as dimensions to present the performance differences of the USAD model compared with mainstream models such as CNN, LSTM, ResNet50, SERes2Net, MAG - Res2Net, ATFA, and TCCSNet on datasets with different degrees of class imbalance. }
    \label{Figure_3}
\end{figure}

For classification accuracy comparison, our research uses a radar chart Figure \ref{Figure_3} to visually compare the performance of different metrics.

\textbf{Mild Imbalance Scenario:} On the WISDM dataset, the USAD model achieves the best performance across all evaluation metrics, with an accuracy of 98.28\%, precision of 97.94\%, recall of 98.39\%, F1-score of 98.16\%, G-mean of 99.05\%, and AUC of 99.01\%, significantly outperforming other baseline models. In contrast, traditional CNN and LSTM models perform poorly in terms of G-mean (67.60\% and 66.63\%, respectively), indicating limited capability in recognizing minority classes. While deeper networks such as ResNet50 and SERes2Net show improvements in accuracy and F1-score, their class balance (G-mean) remains inferior to USAD. Although high-performance models like MAG-Res2Net approach USAD in some metrics, there is still a noticeable gap in overall performance.

\textbf{Moderate Imbalance Scenario:} On the PAMAP2 dataset, the class imbalance issue becomes more pronounced. The USAD model continues to demonstrate strong robustness and generalization, achieving the highest accuracy (94.07\%), precision of 94.34\%, recall of 93.33\%, F1-score (93.72\%), G-mean (96.37\%), and AUC (96.39\%) among all models. In comparison, traditional deep models (such as CNN, LSTM, and ResNet50) show clear deficiencies in G-mean and AUC, which reflect class balance, with CNN’s G-mean as low as 0.7971. Even attention-based models like SERes2Net and MAG-Res2Net lag significantly behind USAD in G-mean and AUC, further validating the superiority of USAD in moderately imbalanced scenarios.

\textbf{Severe Imbalance Scenario:} In the OPPORTUNITY dataset, which represents a severely imbalanced scenario, the advantages of the USAD model become even more pronounced. USAD achieves an accuracy of 84.68\%, precision of 78.47\%, recall of 79.85\%, F1-score of 78.46\%, G-mean of 88.95\%, and AUC of 89.44\%, all of which are substantially higher than those of the comparison models. Traditional models (such as CNN, LSTM, and ResNet50) perform extremely poorly in terms of G-mean and AUC, indicating poor adaptability to extreme class imbalance. Even recent high-performing methods such as MAG-Res2Net, ATFA, and TCCSNet exhibit much lower G-mean and AUC on the OPPORTUNITY dataset compared to USAD, further demonstrating the effectiveness of USAD under extreme class imbalance conditions.

\begin{figure}[htbp]
\centering
\includegraphics[width=0.8\textwidth]{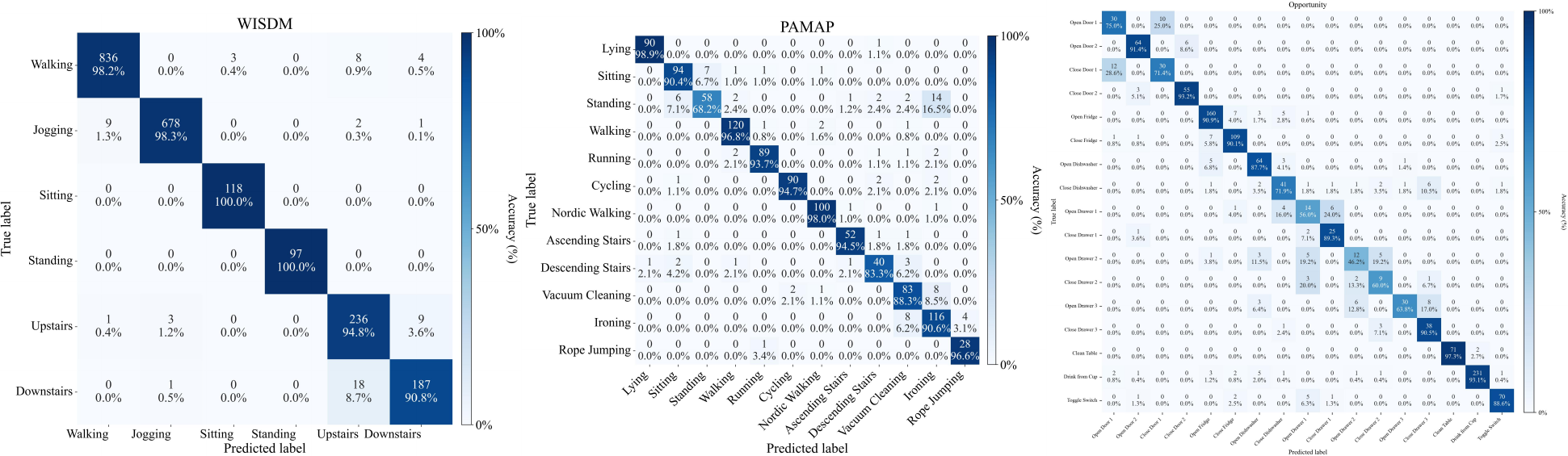}
\caption{Confusion matrices for WISDM, PAMAP2, and OPPORTUNITY.}
\label{Figure_4}
\end{figure}

As shown as Figure in \ref{Figure_4} and \ref{Figure_5},a further analysis of the confusion matrix reveals that USAD maintains a low miss rate and misclassification rate for minority classes, with the most prominent performance observed along the main diagonal. This fully demonstrates its heightened sensitivity to rare activity categories and superior generalization capability.

In summary, the experimental results across the three datasets show that the USAD method consistently exhibits excellent classification performance and strong class balance under varying degrees of class imbalance. Whether in mild, moderate, or severe imbalance scenarios, USAD significantly improves G-mean and AUC—metrics that reflect the model’s ability to recognize minority classes—and consistently leads in overall performance metrics such as accuracy and F1-score. These results fully validate the effectiveness and generalizability of the USAD method in addressing the class imbalance problem.

\subsubsection{T-SNE Visualization Analysis}

\begin{figure}[htbp]
    \centering
    \includegraphics[width=0.9\textwidth]{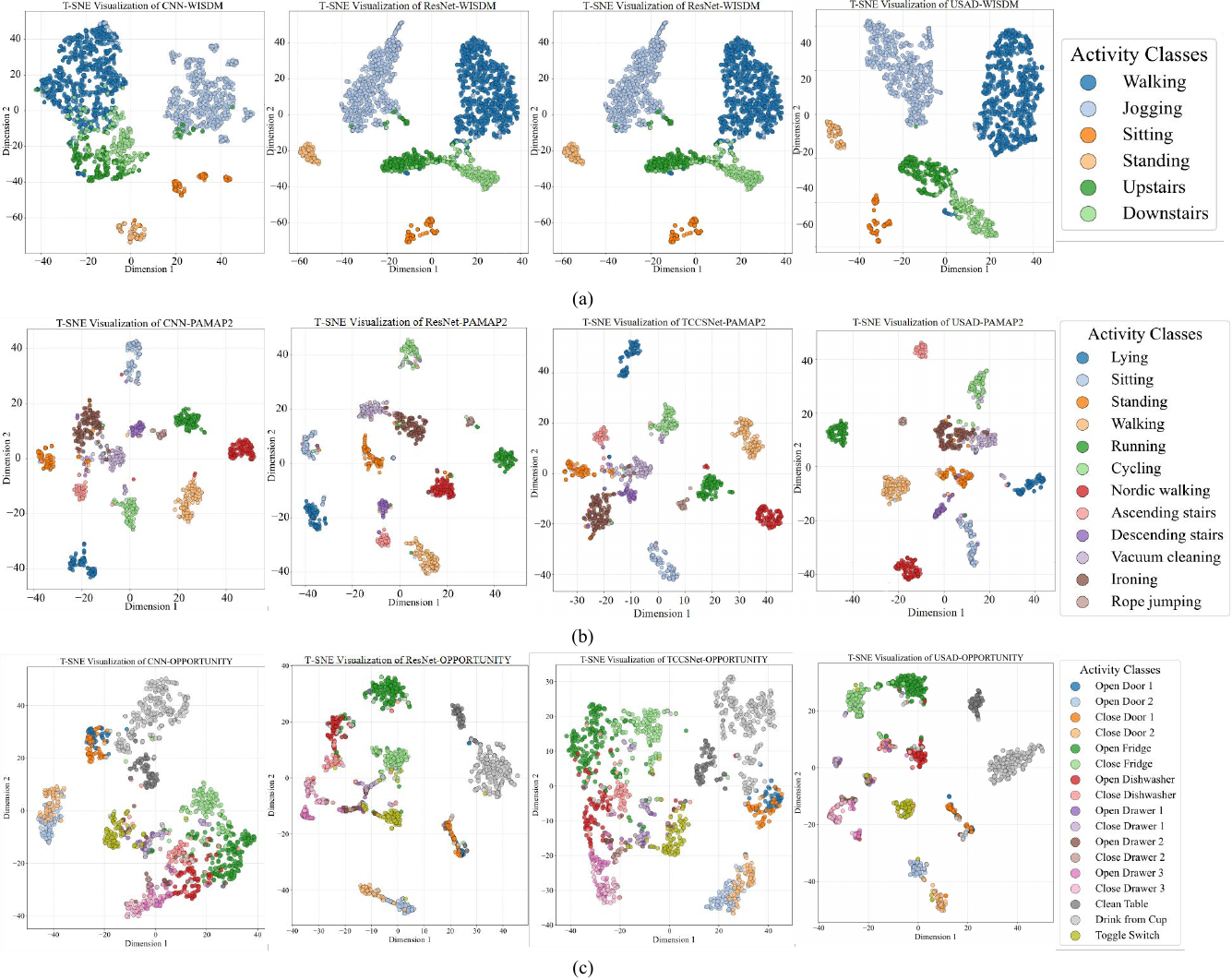}
    \caption{
    T-SNE Visualization Comparison: Feature Distributions of CNN, ResNet50, TCCSNet, and USAD Models on the (a) WISDM, (b) PAMAP2, and (c) OPPORTUNITY Dataset. Results of additional comparison models are provided in the Supplementary Material.
    }
    \label{Figure_5}
\end{figure}

To more intuitively illustrate the feature space distributions of different models, we employed the T-SNE algorithm to reduce the dimensionality of feature vectors from the test samples and visualized them in a two-dimensional space. As shown in Figure \ref{Figure_5}, USAD exhibits outstanding feature separability across all three datasets. Except for a very small number of misclassified samples, USAD enables clear, tightly clustered, and well-separated distributions for different classes, effectively mitigating the adverse impact of class imbalance on the model’s discriminative ability.

In contrast, traditional models (such as CNN, LSTM, and ResNet50) as well as some high-performance existing methods (such as MAG-Res2Net, ATFA, and TCCSNet) generally display blurred class boundaries and severe inter-class overlap in the visualization results. Even ResNet50, which performs relatively well, still shows large intra-class distances on the WISDM dataset, and reduced inter-class distances with more misclassified samples on the OPPORTUNITY datasets.

Notably, USAD also demonstrates highly compact feature distributions for minority classes. For example, for the Rope Jumping class in the PAMAP2 dataset, USAD achieves effective clustering of rare actions, fully representing and recognizing minority categories. In distinguishing similar actions, USAD also shows significant advantages; for instance, for the Ascending/Descending Stairs classes in the PAMAP2 dataset, USAD achieves ideal inter-class separation and intra-class aggregation, whereas models such as SERes2Net fail to reach comparable results.

On the OPPORTUNITY dataset, although USAD exhibits relatively small inter-class distances for categories such as Open/Close 2 and Open/Close Drawer 3, it is still able to effectively separate rare actions from mainstream actions overall, significantly reducing class overlap and providing a solid feature foundation for robust recognition in complex scenarios. However, in extremely challenging cases, such as the Open/Close 1 class in the OPPORTUNITY dataset, USAD still encounters difficulties in discrimination, indicating that there remains room for further improvement in recognizing extremely similar categories.

The experimental results demonstrate that USAD enables new samples to cluster closely within the true class distribution regions in the high-dimensional feature space. Even under class-imbalanced conditions, USAD consistently exhibits excellent overall accuracy, remarkable minority class recognition capability, and outstanding separability of similar features. These advantages provide a solid technical foundation for the deployment and application of human activity recognition systems in real-world complex scenarios.

\subsection{Comparison of Related Work}

We compared our models against a range of methods, including traditional CNN and LSTM network frameworks, as well as models with state-of-the-art performance models. Furthermore, to accurately describe the performance of our method, networks related to our strategy such as SE-Res2Net, Gated-Res2Net, MAGRes2Net were also considered for comparison. We used the WISDM and PAMAP2 data sets for evaluation. Considering the requirements for data enhancement, we adopted a method inspired by \cite{zuo2023unsupervised} and selected the three-axis acceleration data from the PAMAP2 data set for enhancement and testing. The test results are shown in table \ref{Table_10}. 

\begin{table}[htbp]
    \centering
%    \renewcommand{\arraystretch}{1.2}
%    \resizebox{\textwidth}{!}{
        \begin{tabular}{lcccccc}
            \toprule
            \multirow{2}{*}{\textbf{Model}} & \multicolumn{2}{c}{\textbf{PAMAP2 (100\%)}} & \multicolumn{2}{c}{\textbf{PAMAP2 (50\%)}} & \multicolumn{2}{c}{\textbf{WISDM}} \\
            \cmidrule(r){2-3} \cmidrule(r){4-5} \cmidrule(r){6-7}
            & \textbf{Acc. (\%)} & \textbf{F1 (\%)} & \textbf{Acc. (\%)} & \textbf{F1 (\%)} & \textbf{Acc. (\%)} & \textbf{F1 (\%)} \\
            \midrule
            CNN\cite{zeng2014convolutional} & 83.80 & 79.89 & 60.29 & 58.25 & 91.09 & 90.01 \\
            LSTM\cite{xia2020lstm} & 82.38 & 80.29 & 55.10 & 47.37 & 80.59 & 75.67 \\
            LSTM-CNN\cite{xia2020lstm} & 57.71 & 53.31 & 52.57 & 48.68 & 95.90 & 95.97 \\
            CNN-GRU\cite{dua2021multi} & 61.43 & 54.92 & 60.48 & 52.15 & 94.95 & 96.21 \\
            SE-Res2Net\cite{gao2019res2net} & 88.04 & 82.20 & 66.37 & 62.46 & 98.23 & 97.71 \\
            ResNeXt\cite{mekruksavanich2022deep} & 62.48 & 60.02 & 58.57 & 53.77 & 96.67 & 96.66 \\
            Gated-Res2Net\cite{yang2020gated} & 70.48 & 68.75 & 65.52 & 64.31 & 97.02 & 97.02 \\
            Rev-Attention\cite{pramanik2023transformer} & 63.90 & 61.59 & 58.86 & 57.37 & 97.46 & 97.49 \\
            MAG-Res2Net\cite{liu2023mag} & 88.57 & 87.80 & 67.34 & 65.87 & 97.24 & 97.00 \\
            HAR-CE\cite{CE} & 65.90 & 58.55 & 61.81 & 58.35 & 97.72 & 97.70 \\
            ELK\cite{ELK} & 72.95 & 72.77 & \textbf{72.57} & \textbf{72.42} & 98.05 & 98.06 \\
            DanHAR\cite{DanHAR} & 71.52 & 70.27 & 70.57 & 68.18 & 95.57 & 95.64 \\
            \textbf{USAD} & \textbf{94.07} & \textbf{93.72} & 89.29 & 88.92 & \textbf{98.84} & \textbf{98.79} \\
            \bottomrule
        \end{tabular}
%    }
    \caption{Comparison of related work}
    \label{Table_10}
\end{table}

\begin{table}[htbp]
    \centering
    \label{tab:WISDM-Test}
%    \resizebox{\textwidth}{!}{
        \begin{tabular}{l@{\hspace{0.1cm}}cccc@{\hspace{0.1cm}}cccc@{\hspace{0.1cm}}cccc}
            \toprule
            \multirow{2}{*}{\textbf{Activity}} & \multicolumn{2}{c}{\textbf{LSTM-CNN}} & \multicolumn{2}{c}{\textbf{CNN-GRU}} & \multicolumn{2}{c}{\textbf{SE-Res2Net}} & \multicolumn{2}{c}{\textbf{ResNeXt}} & \multicolumn{2}{c}{\textbf{Gated-Res2Net}} \\
            \cmidrule(r){2-3} \cmidrule(r){4-5} \cmidrule(r){6-7} \cmidrule(r){8-9} \cmidrule(r){10-11}
            & \textbf{Acc.} & \textbf{F1} & \textbf{Acc.} & \textbf{F1} & \textbf{Acc.} & \textbf{F1} & \textbf{Acc.} & \textbf{F1} & \textbf{Acc.} & \textbf{F1} \\
            \midrule
            Downstairs & 89.29 & 90.99 & 90.37 & 93.93 & 80.16 & 88.56 & 90.43 & 86.77 & 87.14 & 89.98 \\
            Jogging & 98.96 & 99.25 & 99.19 & 99.31 & 99.28 & 99.30 & 99.08 & 99.02 & 98.86 & 99.17 \\
            Sitting & 98.31 & 98.98 & 98.30 & 99.49 & 98.31 & 98.32 & 99.32 & 98.82 & \textbf{99.66} & 98.39 \\
            Standing & 98.15 & 97.15 & 98.56 & 97.56 & 98.13 & 98.34 & 97.95 & 98.45 & 97.56 & 98.56 \\
            Upstairs & 90.76 & 91.07 & 91.36 & 89.46 & 96.43 & 89.13 & 80.44 & 87.71 & 86.42 & 91.00 \\
            Walking & 98.49 & 97.77 & 98.46 & 98.06 & 98.81 & 98.92 & 99.52 & 98.62 & 99.27 & 99.16 \\
            \midrule
            \multirow{2}{*}{\textbf{Activity}} & \multicolumn{2}{c}{\textbf{Rev-Attn}} & \multicolumn{2}{c}{\textbf{MAG-Res2Net}} & \multicolumn{2}{c}{\textbf{HAR-CE}} & \multicolumn{2}{c}{\textbf{ELK}} & \multicolumn{2}{c}{\textbf{DanHAR}} & \multicolumn{2}{c}{\textbf{USAD}} \\
            \cmidrule(r){2-3} \cmidrule(r){4-5} \cmidrule(r){6-7} \cmidrule(r){8-9} \cmidrule(r){10-11} \cmidrule(r){12-13}
            & \textbf{Acc.} & \textbf{F1} & \textbf{Acc.} & \textbf{F1} & \textbf{Acc.} & \textbf{F1} & \textbf{Acc.} & \textbf{F1} & \textbf{Acc.} & \textbf{F1} & \textbf{Acc.} & \textbf{F1} \\
            \midrule
            Downstairs & 88.54 & 92.46 & 91.25 & 91.46 & 91.46 & 93.28 & 91.85 & 93.19 & 92.60 & 85.99 & \textbf{95.86} & \textbf{95.91} \\
            Jogging & 99.20 & 99.14 & 99.20 & 98.87 & 99.42 & 99.32 & 99.51 & 99.19 & 97.85 & 98.65 & \textbf{99.60} & \textbf{99.42} \\
            Sitting & 99.15 & 99.07 & 98.33 & 98.56 & 99.49 & 98.73 & 99.66 & 99.16 & 99.83 & 99.07 & 99.16 & \textbf{99.08} \\
            Standing & 98.34 & 98.47 & \textbf{98.74} & 98.27 & 97.36 & 98.46 & 97.96 & 98.77 & 98.36 & 99.07 & 97.97 & \textbf{98.97} \\
            Upstairs & 95.44 & 92.36 & 93.46 & 91.23 & \textbf{97.09} & 92.50 & 94.57 & 94.04 & 77.23 & 85.35 & 95.72 & \textbf{95.56} \\
            Walking & 99.34 & 99.33 & 98.71 & 98.64 & 97.84 & 98.64 & 99.13 & 99.16 & \textbf{99.83} & 97.46 & \textbf{99.36} & 99.43 \\
            \bottomrule
        \end{tabular}
%    }
    \caption{Classification results of the WISDM dataset}
    \label{Table_11}
\end{table}

Notably, as shown as Table \ref{Table_11}, USAD demonstrates higher accuracy on the PAMAP2 dataset with 100\% data volume. On PAMAP2 with 50\% of the data volume, the comprehensive performance of USAD has also reached an extremely high level, surpassing all models except ELK. USAD achieves an accuracy of 98. 84\% on WISDM, exceeding the highest benchmark previously set by other models. These results verify the superior performance of our model on both simple and complex datasets. 

\subsection{Actual Deployment}

We designed four complexity index test experiments on the WISDM and OPPO data sets for the baseline and SOTA models, that is, the model is in the process of inferring each data set. model parameter size, memory consumption, inference latency, and inference latency of a single fragment. Since PyTorch can currently support the deployment of deep learning models on Raspberry Pi 5, this work was performed on Raspberry Pi 5 equipped with a quad-core Cortex-A76  (ARM) 64-bit SoC @ 2.4 GHz and 8gb LPDDR4X-4267 SDRAM, and the pytorch version is 2.2.2. Figure \ref{Figure_1} shows our software interfaceFor reference, here are some examples of smartphone chips with comparable overall performance to this processor: Snapdragon 865/Apple A12/Exynos 1080/Dimensity 1200 or Kirin 9000SL. 

\begin{table}[h]
\centering
%\scalebox{0.9}{
\begin{tabular}{c|c|c|c|c}
\toprule
\multirow{2}{*}{\textbf{Model}} & \multicolumn{2}{c|}{\textbf{WISDM}} & \multicolumn{2}{c}{\textbf{OPPORTUNITY}} \\
\cmidrule(r){2-3} \cmidrule(r){4-5}
 & Memory & Param. & Memory & Param. \\
\cmidrule(r){1-1} \cmidrule(r){2-3} \cmidrule(r){4-5}
CNN & 766.23 & 1.04E+05 & 744.52 & 1.21E+05 \\
\cmidrule(r){1-1} \cmidrule(r){2-3} \cmidrule(r){4-5}
LSTM & 759.80 & 1.65E+05 & 748.44 & 3.85E+05 \\
\cmidrule(r){1-1} \cmidrule(r){2-3} \cmidrule(r){4-5}
CNN-GRU & 787.05 & 4.38E+06 & 773.17 & 4.39E+06 \\
\cmidrule(r){1-1} \cmidrule(r){2-3} \cmidrule(r){4-5}
LSTM-CNN & 823.00 & 2.12E+06 & 788.69 & 2.13E+06 \\
\cmidrule(r){1-1} \cmidrule(r){2-3} \cmidrule(r){4-5}
ResNeXt & 889.78 & 2.20E+07 & 849.25 & 2.21E+07 \\
SE-Res2Net & 787.67 & 1.60E+06 & 772.41 & 1.61E+06 \\
\cmidrule(r){1-1} \cmidrule(r){2-3} \cmidrule(r){4-5}
Rev-Att & 862.03 & 4.33E+05 & 801.97 & 3.80E+05 \\
\cmidrule(r){1-1} \cmidrule(r){2-3} \cmidrule(r){4-5}
Gated-Res2Net & 773.84 & 1.60E+06 & 769.53 & 1.61E+06 \\
\cmidrule(r){1-1} \cmidrule(r){2-3} \cmidrule(r){4-5}
HAR-CE & 788.73 & 4.17E+05 & 779.84 & 3.49E+06 \\
\cmidrule(r){1-1} \cmidrule(r){2-3} \cmidrule(r){4-5}
ELK & 782.70 & 2.41E+05 & 751.84 & 7.57E+05 \\
\cmidrule(r){1-1} \cmidrule(r){2-3} \cmidrule(r){4-5}
DanHAR & 790.66 & 2.35E+06 & 814.09 & 4.46E+06 \\
\cmidrule(r){1-1} \cmidrule(r){2-3} \cmidrule(r){4-5}
USAD & 718.50 & 3.47E+05 & 751.30 & 1.09E+06 \\
\bottomrule
\end{tabular}
%}
\caption{Complexity analysis. }
\label{Table_12}
\end{table}

In order to ensure the stability of the test, we continuously tested all models 100 times. The evaluated indicators include running time, memory usage. We did not record the results of running these models on the entire data set, because most of them were unable to complete this test, and we believe that such a test is of no practical value without adapting to the memory management method. Table \ref{Table_12} shows the average results of ten rounds of 100 tests on WISDM and OPPO datasets. It can be clearly seen that the memory consumption of our proposed method is lower than most state-of-the-art work, while the model parameters are fewer among them.

In order to further explore the absolute efficiency of the model, we set up an inference delay test to test the time it takes for the model to infer a single sample and continuously infer 100 samples. We referred to Zhang's inference delay test method and tested the sensor data through a sliding step window \cite{NNU2023}. They selected a 10s action clip , the step size of the sliding window is 95\% of the total length of the window, which is 500ms, and it is considered that less than 500ms meets the practical standard. However, after our testing, we found that due to the improvement of the performance of the deployment equipment, most models can easily achieve this standard. However, for the sliding window applicable to the public data set, compared with the commonly used window length of the data set, the previously selected 10s is a very huge value. For example, the commonly used segment length in WISDM is 4s, which is the most commonly used sliding window length in general. Then the corresponding window step size is 200ms, which means that the system will segment the data and predict every 200 milliseconds. New sample. We apply this approach to each dataset, i.e. the segments are segmented exactly as in the base experiment, and the latency requirement for model inference for a single segment is set to a 5\% window of the length of the individual segment to ensure the model is segmented in the next segment.

\begin{figure}[ht]
    \centering
%\captionsetup[figure]{justification=raggedright}
    \begin{minipage}{0.48\textwidth}
        \includegraphics[width=\linewidth]{Figure_10.pdf}
        \caption{Inference delay of each model on the WISDM dataset}
        \label{Figure_10}
    \end{minipage}
    \hfill
    \begin{minipage}{0.48\textwidth}
        \includegraphics[width=\linewidth]{Figure_11.pdf}
        \caption{Inference delay of each model on the PAMAP2 dataset}
        \label{Figure_11}
    \end{minipage}
\end{figure}

%\begin{figure}[ht]
%    \centering
%    \includegraphics[width=0.7\linewidth]{Figure_10.pdf}
%    \caption{Inference delay of each model on the WISDM dataset}
%    \label{Figure_10}
%\end{figure}
%
%\begin{figure}[ht]
%    \centering
%    \includegraphics[width=0.7\linewidth]{Figure_11.pdf}
%    \caption{Inference delay of each model on the PAMAP2 dataset}
%    \label{Figure_11}
%\end{figure}

The experimental results are shown in Figure  \ref{Figure_10} and Figure  \ref{Figure_11}. On the WISDM data set, all the models we tested can complete the task within the limited time, and most of the models are concentrated in the lower time interval. This is consistent with our expectations for the performance of state-of-the-art embedded devices and the complexity of data sets. On the OPPORTUNITY data set, there is a clear gap in the efficiency of each model. The basic model still has a lower time, and some lightweight models, such as HAR-CE, ELK and our USAD-AD can also be maintained at Within 25ms, while ResNeXt completely exceeded the time limit.

\section{Conclusion}

This study proposes a novel HAR framework named USAD, which aims to address core challenges such as data scarcity, class imbalance, and insufficient feature extraction. USAD significantly enhances model performance by introducing an unsupervised diffusion model for data augmentation, designing a multi-branch spatio-temporal interaction network to capture multi-scale features and integrate attention mechanisms, and employing an adaptive composite loss function to optimize the training process. Experimental results demonstrate that USAD achieves outstanding classification accuracy, minority class recognition, and generalization ability across multiple public datasets. Furthermore, this study verifies the efficient deployment of USAD on resource-constrained embedded devices, providing a practical and robust solution for real-world HAR applications.

\bibliographystyle{elsarticle-num}

\bibliography{refs}

\begin{appendices}

\section{Experimental Design}

\subsection{Datasets}

All model training and deployment experiments were conducted under two distinct computational environments. The training phase was performed on a high-performance server equipped with PyTorch 2.5.1 and Python 3.10, dual NVIDIA RTX 4090 (24GB) GPUs, 16-core Intel Xeon Gold 6430 processors, and 120GB memory, with CUDA 12.1 support. The deployment phase was executed on a resource-constrained embedded device—a quad-core 64-bit Arm Cortex-A76 CPU with 8GB of memory, running PyTorch 2.2.2 and Python 3.11.2, without GPU or CUDA acceleration.

For performance evaluation, three widely used human activity recognition datasets were selected. Prior to input into the USAD model, each dataset was preprocessed and parameterized according to their specific properties and experimental protocol, as detailed below.

\textbf{WISDM} \cite{kwapisz2011activity}: The WISDM dataset comprises activity records from 29 participants performing six common daily activities: walking, jogging, sitting, standing, upstairs and downstairs. Data were collected using three tri-axial accelerometers worn on the wrist. For experiments with the USAD model, the raw signals were segmented using a sliding window of size 90, with a batch size of 512. The model was trained for 50 epochs with a learning rate of 0.005.

\textbf{PAMAP2} \cite{reiss2012introducing}: The PAMAP2 dataset contains sensor data from nine subjects engaged in 12 different physical activities. Data acquisition involved three types of sensors—accelerometers, gyroscopes, and magnetometers—placed on the wrist, chest, and ankle, resulting in 40 sensor channels. For model training, a sliding window of size 171 was applied, with a batch size of 256 and an initial learning rate of 0.0005. The model was trained for 70 epochs. The number of activity classes and subjects follows the original dataset specifications.

\textbf{OPPORTUNITY}\cite{R.Chavarriaga}: The OPPORTUNITY dataset offers sensor-rich recordings of 12 participants engaged in naturalistic activities within a sensor-enhanced environment. It integrates 72 sensors across 17 types, providing comprehensive multimodal data for human activity recognition research. For experimental consistency, preprocessing applied a window size of 30, batch size of 128, learning rate of 0.001, and 80 epochs of training.

\subsection{Evaluation Metrics}

To comprehensively evaluate the performance of the proposed HAR model, we employ multiple evaluation metrics, each capturing a distinct aspect of model effectiveness. Accuracy (Acc.) measures the proportion of correctly classified samples among all predictions, providing an overall indicator of model performance. Precision (Pre.) quantifies the proportion of correctly predicted positive samples among all samples predicted as positive, while Recall (Rec.) measures the proportion of correctly predicted positive samples among all actual positive samples. F1-weighted (F1-wt.) is the weighted average of F1 scores across all classes, accounting for class imbalance, whereas F1-macro (F1-mac.) computes the unweighted mean F1 score, treating each class equally regardless of its frequency. Geometric Mean (G-mean) evaluates the balance between sensitivity and specificity, highlighting the model’s ability to maintain robust recognition across both majority and minority classes, which is particularly important when training on augmented data. Additionally, we report the Area Under the Receiver Operating Characteristic Curve (AUC), which quantifies the model’s ability to distinguish between classes across various threshold settings and is especially informative for imbalanced datasets. The combined use of these metrics enables a comprehensive and nuanced analysis of the model’s strengths and limitations in HAR tasks.
\begin{flushleft}
\begin{align*}
\text{Accuracy (Acc.)} &= \frac{\mathrm{TP} + \mathrm{TN}}{\mathrm{TP} + \mathrm{FN} + \mathrm{FP} + \mathrm{TN}} \\
\text{Precision (Pre.)}_i &= \frac{\mathrm{TP}_i}{\mathrm{TP}_i + \mathrm{FP}_i} \\
\text{Recall (Rec.)}_i &= \frac{\mathrm{TP}_i}{\mathrm{TP}_i + \mathrm{FN}_i} \\
\text{F1-macro (F1-mac.)} &= \frac{1}{N} \sum_{i=1}^{N} \frac{2 \times \text{Precision}_i \times \text{Recall}_i}{\text{Precision}_i + \text{Recall}_i} \\
\text{F1-weighted (F1-wt.)} &= \sum_{i=1}^{N} \omega_i \times \frac{2 \times \text{Precision}_i \times \text{Recall}_i}{\text{Precision}_i + \text{Recall}_i} \\
\text{G-mean} &= \sqrt{\frac{\mathrm{TP}}{\mathrm{TP} + \mathrm{FN}} \times \frac{\mathrm{TN}}{\mathrm{TN} + \mathrm{FP}}} \\
\text{AUC} &= \int_{0}^{1} \text{TPR}(t) \, d\text{FPR}(t) \\
&\approx \sum_{k=1}^{K-1} \frac{[\text{TPR}(k+1) + \text{TPR}(k)]}{2} \times [\text{FPR}(k+1) - \text{FPR}(k)]
\end{align*}
\end{flushleft}

where $\mathrm{TP}$ and $\mathrm{TN}$ represent the numbers of true positives and true negatives, respectively, while $\mathrm{FP}$ and $\mathrm{FN}$ denote the numbers of false positives and false negatives. $\mathrm{Precision}_i$ and $\mathrm{Recall}_i$ are the precision and recall for class $i$, and $\omega_i$ is the proportion of samples in class $i$. $N$ is the total number of classes. AUC (Area Under the Curve) is calculated as the area under the ROC curve, where TPR and FPR denote the true positive rate and false positive rate, respectively.

\section{Analysis of Feature Discriminability}
To further investigate the performance of different models in the feature space, we calculated the intra-class distance, inter-class distance, and their ratio (Inter/Intra ratio) on the test set for each model. As shown in Table \ref{Table_2}, USAD demonstrates highly competitive feature discriminability across all three datasets.

\begin{table}[htbp]
    \centering
    \label{tab:distance_analysis}
%    \resizebox{\textwidth}{!}{
    \begin{tabular}{lccccccccc}
        \toprule
        \multirow{2}{*}{Model} & \multicolumn{3}{c}{\textbf{WISDM}} & \multicolumn{3}{c}{\textbf{PAMAP2}} & \multicolumn{3}{c}{\textbf{OPPORTUNITY}} \\
        \cmidrule(r){2-4} \cmidrule(r){5-7} \cmidrule(r){8-10}
        & Intra & Inter & Ratio & Intra & Inter & Ratio & Intra & Inter & Ratio \\
        \midrule
        CNN & 2.525 & 6.652 & 2.635 & 2.816 & 7.875 & 2.797 & 4.777 & 7.957 & 1.666 \\
        LSTM & 1.047 & 6.242 & 5.961 & 1.726 & 6.883 & 3.988 & \textbf{2.043} & 6.449 & 3.156 \\
        ResNet50 & 0.446 & 6.508 & 14.594 & 1.583 & 6.930 & 4.379 & 3.225 & 7.774 & 2.410 \\
        SERes2Net & 1.320 & 6.598 & 4.996 & \textbf{1.491} & 7.102 & 4.762 & 2.668 & 6.727 & 2.522 \\
        MAG-Res2Net & 0.490 & 6.543 & 13.344 & 1.885 & 6.824 & 3.621 & 3.080 & 6.648 & 2.158 \\
        ATFA & 2.797 & 6.930 & 2.478 & 3.007 & \textbf{9.907} & 3.294 & 4.850 & 8.379 & 1.728 \\
        TCCSNet & 2.906 & 6.930 & 2.385 & 3.018 & 8.289 & 2.746 & 4.77 & \textbf{8.492} & 1.779 \\
        USAD & \textbf{0.386} & \textbf{6.578} & \textbf{17.063} & 1.670 & 7.336 & \textbf{4.395} & 2.170 & 7.098 & \textbf{3.271} \\
        \bottomrule
    \end{tabular}
%    }
    \caption{Intra-class/Inter-class Distances and Feature Separability(Intra/Inter Ratio) in the Feature Space}
    \label{Table_2}
\end{table}

Specifically, on the WISDM dataset, USAD achieves the optimal Inter/Intra ratio, with the smallest intra-class distance and the largest inter-class distance, indicating that its deep modeling strategy greatly enhances the distinguishability of rare classes. On the more challenging PAMAP2 and OPPORTUNITY datasets, USAD also achieves the best feature separation ratio, and its individual metrics (intra-class and inter-class distances) outperform those of most comparison models. These results further demonstrate that USAD not only exhibits excellent feature aggregation and separation capabilities in simple scenarios, but also ensures accurate recognition of minority classes and maintains a well-structured feature space in complex tasks.

\section{Hyperparameter Exploration of the Adaptive Compound Loss Function on the PAMAP2 Dataset}

To further enhance the robustness, generalization, and calibration reliability of the USAD method in human activity recognition tasks, we systematically conducted experimental studies on the core hyperparameters of the adaptive compound loss function. The overall experimental process is divided into three stages: single loss optimization, composite weight tuning, and sensitivity analysis.

\subsection{Independent Optimization of Single Loss Function Hyperparameters}

Firstly, we performed hyperparameter optimization for three fundamental loss functions—Label Smoothing, Focal Loss, and Class Balanced Loss (CB Loss)—to ensure that each loss function achieves optimal performance when used individually.

\begin{figure}[htbp]
    \centering
\includegraphics[width=\textwidth]{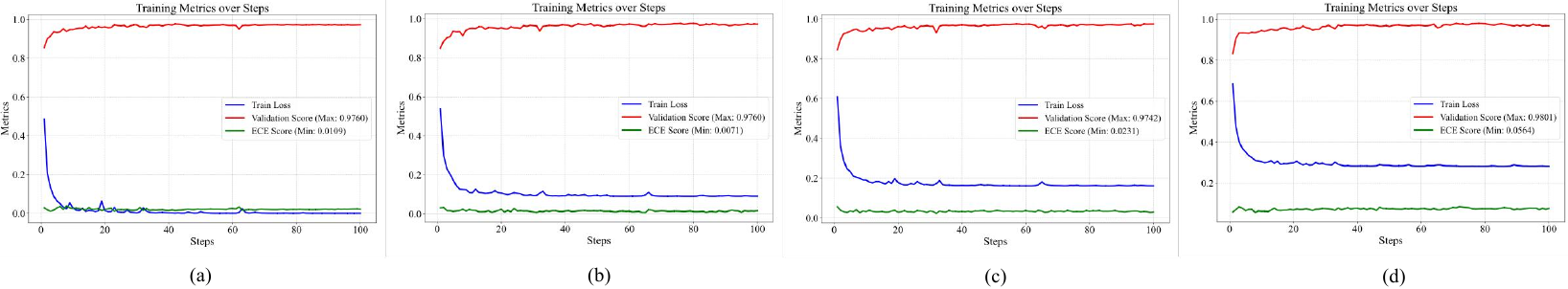}
    \caption{Training dynamics under label smoothing. 
    \textbf{(a) $\text{$\hat{y}$}=0.05$}: Optimal balance between validation accuracy (0.9760) and ECE (0.0109). 
    \textbf{(b) $\text{$\hat{y}$}=0.1$}: Higher ECE (0.0125) due to over-smoothing, despite comparable accuracy. 
    \textbf{(c) $\text{$\hat{y}$}=0.01$}: Lower accuracy (0.9653) and slower convergence. 
    \textbf{(d) $\text{$\hat{y}$}=0.2$}: Drastic accuracy drop (0.9411) and unstable training. 
    These configurations highlight the sensitivity of model calibration to smoothing coefficients.}
    \label{Figure_6}
\end{figure}
  
\textbf{Label Smoothing:} The smoothing coefficient ($\hat{y}$) regulates label distribution "softening". As shown in Table \ref{Table_3}, a moderate value balances classification accuracy and calibration: it minimizes the expected calibration error (ECE, a metric for consistency between predicted probabilities and actual accuracies) while maintaining performance. As visualized in Figure \ref{Figure_6}, with $\hat{y}$ = 0.05, training loss drops steadily, validation scores stay high (e.g., 0.9760 in subplot (a)), and ECE is minimized (e.g., 0.0109 in subplot (a)). This shows well-calibrated predictions and strong classification ability. Excessive/low coefficients harm discriminative power (over-softened labels blur boundaries or fail to regularize confidence). Therefore, it is recommended to select a relatively small smoothing coefficient in practical applications to achieve the best overall results.

\begin{table}[htbp]
\centering
\begin{minipage}[t]{0.32\textwidth}
    \centering
\vspace{0pt}
    \resizebox{\textwidth}{!}{
        \begin{tabular}{lccc}
            \toprule
            \textbf{$\hat{y}$} & \textbf{Acc. (\%)} & \textbf{F1 (\%)} & \textbf{ECE} \\
            \midrule
            0 & \textbf{85.90} & \textbf{85.68} & 8.36 \\
            0.05 & 85.81 & 85.58 & \textbf{5.74} \\
            0.1 & 80.19 & 79.12 & 5.58 \\
            0.2 & 82.95 & 82.02 & 8.50 \\
            \bottomrule
        \end{tabular}
    }
    \caption{Sensitivity analysis of label smoothing\\ coefficients.}
    \label{Table_3}
\end{minipage}
\hfill
\begin{minipage}[t]{0.32\textwidth}
    \centering
\vspace{0pt}
    \resizebox{\textwidth}{!}{
        \begin{tabular}{cccc}
            \toprule
            $\gamma$ & $\alpha$ & \textbf{Acc. (\%)} & \textbf{F1 (\%)} \\
            \midrule
            1 & 0.25 & \textbf{92.69} & \textbf{92.50} \\
            2 & 0.25 & 91.68 & 91.35 \\
            3 & 0.25 & 89.74 & 88.62 \\
            1 & 0.5 & 91.84 & 91.67 \\
            1 & 0.75 & 92.38 & 92.15 \\
            \bottomrule
        \end{tabular}
    }
    \caption{Focal loss hyperparameter tuning.}
    \label{Table_4}
\end{minipage}
\hfill
\begin{minipage}[t]{0.32\textwidth}
    \centering
\vspace{0pt}
    \resizebox{\textwidth}{!}{
        \begin{tabular}{lcc}
            \toprule
            $\beta$ & \textbf{Acc. (\%)} & \textbf{F1 (\%)} \\
            \midrule
            0.9 & \textbf{93.00} & \textbf{92.99} \\
            0.99 & 90.82 & 90.52 \\
            0.999 & 89.50 & 88.88 \\
            \bottomrule
        \end{tabular}
    }
    \caption{Class-balanced loss parameter optimization.}
    \label{Table_5}
\end{minipage}
\end{table}

\textbf{Focal Loss:} As shown in Table \ref{Table_4}, focal Loss introduces the gamma ($\gamma$) parameter to dynamically adjust the focus on easy and hard samples, thereby enhancing the model’s ability to distinguish between similar actions. The alpha ($\alpha$)  parameter is used to balance the loss weights of different classes, alleviating the class imbalance problem and further improving the recognition of minority classes. As shown in the table, when gamma increases from 1 to 3, both the model’s accuracy and F1 score gradually decrease, indicating that an excessively high focusing factor weakens the model’s learning ability for majority samples and adversely affects overall performance. When gamma is set to 1, appropriately increasing alpha helps improve the model’s recognition of minority classes. Experimental results demonstrate that the combination of $\gamma$ = 1 and $\alpha$ = 0.25 achieves the best performance in scenarios with class imbalance and mixed sample difficulty, while an excessively high gamma leads to unstable training.

\textbf{Class Balanced Loss (CB Loss): } As shown in Table \ref{Table_5}, the beta ($\beta$) coefficient in CB Loss is used to quantify the impact of class sample size; the closer its value is to 1, the greater the compensation for classes with fewer samples, which facilitates feature learning for minority classes and enhances recognition fairness. Experimental results show that when beta is set to 0.9, the performance between majority and minority classes is relatively balanced; however, an excessively high beta leads to a decline in the accuracy of majority classes.

\subsection{Experiments on Composite Loss Weights and Regulation Parameters}

Based on fixing the optimal hyperparameters for each basic loss function, we further investigate the performance of the composite loss under different mechanisms, including weight allocation (initial weights), weight adjustment temperature (temperature), and weight constraints (min weight/max weight). As shown in Table \ref{Table_6}, the main conclusions are as follows:

\textbf{Weight Allocation:} Uniform weighting ([0.33, 0.33, 0.34]) generally achieves the best overall performance (Accuracy: 0.9316, F1: 0.9306) and yields a lower ECE, demonstrating the advantage of multi-loss fusion. Biased weighting (e.g., excessively high weight for a single loss) leads to performance degradation in certain classes, confirming the importance of balanced integration of multiple supervisory signals.

\textbf{Temperature Parameter:} The temperature parameter has a significant impact on the dynamic weight adjustment mechanism. With a moderate temperature (e.g., 1.0), the model achieves optimal generalization. Excessively high or low temperature values tend to result in extreme or overly uniform weight distributions, which impairs the model’s targeted learning capability.

\textbf{Weight Constraint Boundaries:} Appropriately setting weight boundaries (e.g., min = 0.1, max = 0.8) can prevent any individual loss from being excessively ignored or overemphasized, thereby enhancing the robustness of the model.

\begin{table}[htbp]
    \centering
\setlength{\tabcolsep}{2pt}
    \label{tab:hyperparameter_comparison}
%    \resizebox{\textwidth}{!}{
        \begin{tabular}{lcccccccc}
            \toprule
            \textbf{Initial Weights} & \textbf{Temp} & \textbf{Min Weight} & \textbf{Max Weight} & \textbf{Acc. (\%)} & \textbf{F1 (\%)} & \textbf{Pre. (\%)} & \textbf{Rec. (\%)} & \textbf{Remarks} \\
            \midrule
            {[}0.33, 0.33, 0.34{]} & 1 & 0.1 & 0.8 & \textbf{93.16} & \textbf{93.06} & \textbf{92.03} & \textbf{91.85} & Best Performance \\
            {[}0.6, 0.2, 0.2{]} & 1 & 0.1 & 0.8 & 84.19 & 83.78 & 83.96 & 85.60 & Weight Bias \\
            {[}0.2, 0.6, 0.2{]} & 1 & 0.1 & 0.8 & 78.10 & 75.94 & 83.49 & 82.14 & Weight Bias \\
            {[}0.33, 0.33, 0.34{]} & 0.5 & 0.1 & 0.8 & 79.81 & 76.83 & 81.12 & 80.26 & Low Temperature \\
            {[}0.33, 0.33, 0.34{]} & 2 & 0.1 & 0.8 & 78.19 & 75.87 & 78.45 & 79.43 & High Temperature \\
            {[}0.33, 0.33, 0.34{]} & 1 & 0.3 & 0.6 & 78.10 & 76.36 & 79.06 & 78.87 & Narrow Weight Range \\
            {[}0.33, 0.33, 0.34{]} & 1 & 0.05 & 0.9 & 78.19 & 75.87 & 78.45 & 79.43 & Wide Weight Range \\
            \bottomrule
        \end{tabular}
%    }
    \caption{Performance Comparison with Different Hyperparameters}
     \label{Table_6}
\end{table}

\subsection{Sensitivity Analysis and Interleaved Experiments}
To further explore the optimal configuration of loss function hyperparameters under joint optimization, this study conducted sensitivity analysis and interleaved experiments based on the independent optimization of single loss function hyperparameters. Specifically, by alternately adjusting the hyperparameters of different loss functions, we systematically evaluated the interactions among hyperparameters and their combined impact on model performance.

Experimental results indicate that the optimal values of the hyperparameters for the three types of loss functions in the interleaved experiments differ from those obtained during independent optimization. For example, as shown in Table \ref{Table_6}, the optimal $\hat{y}$ was 0.05 in the independent optimization stage, but increased to 0.1 in the interleaved experiments. This phenomenon suggests that, under the synergistic effect of multiple loss functions, the model’s demand for label smoothing is enhanced, and moderately increasing the smoothing coefficient further improves the model’s generalization ability and confidence calibration.

As shown in Table \ref{Table_8}, for Focal Loss, the combination of $\gamma$ = 1 and $\alpha$ = 0.25 yields the best performance during independent optimization, whereas in the interleaved experiments, $\gamma$ = 1 and $\alpha$ = 0.5 become the optimal configuration. This change indicates that, under joint optimization with multiple loss functions, the model’s sensitivity to class imbalance is enhanced. Appropriately increasing the class balancing factor alpha can more effectively improve the recognition of minority classes and optimize overall performance.

As shown in Table \ref{Table_9}, for F-beta Loss, the optimal value of the $\beta$ increases from 0.9 in the independent optimization stage to 0.999 in the interleaved experiments. This result indicates that, under the combined effect of multiple loss functions, the model’s emphasis on recall is further strengthened, and a higher beta value helps to enhance the model’s recall capability for hard-to-detect samples.

\begin{table}[htbp]
\centering
\begin{minipage}[t]{0.32\textwidth}
    \centering
\vspace{0pt}
    \resizebox{\textwidth}{!}{
        \begin{tabular}{lcc}
            \toprule
            \textbf{$\hat{y}$} & \textbf{Accuracy (\%)} & \textbf{F1 Score (\%)} \\
            \midrule
            0.0 & 86.19 & 85.68 \\
            0.05 & 75.33 & 73.17 \\
            0.1 & \textbf{86.67} & \textbf{86.66} \\
            0.2 & 78.29 & 77.13 \\
            \bottomrule
        \end{tabular}
    }
    \captionof{table}{$\hat{y}$ - Model Performance \\Relationship}
    \label{Table_7}
\end{minipage}
\hfill
\begin{minipage}[t]{0.32\textwidth}
    \centering\vspace{0pt}
    \resizebox{\textwidth}{!}{
        \begin{tabular}{cccc}
            \toprule
            $\gamma$ & $\alpha$ & \textbf{Accuracy (\%)} & \textbf{F1 Score (\%)} \\
            \midrule
            1 & 0.25 & 90.86 & 90.75 \\
            2 & 0.25 & 82.48 & 80.68 \\
            3 & 0.25 & 77.43 & 74.38 \\
            1 & 0.5 & \textbf{91.05} & \textbf{90.65} \\
            1 & 0.75 & 87.71 & 87.37 \\
            \bottomrule
        \end{tabular}
    }
    \captionof{table}{$\gamma$ and $\alpha$ - Model Performance Relationship}
    \label{Table_8}
\end{minipage}
\hfill
\begin{minipage}[t]{0.32\textwidth}
    \centering\vspace{0pt}
    \resizebox{\textwidth}{!}{
        \begin{tabular}{lcc}
            \toprule
            $\beta$ & \textbf{Accuracy (\%)} & \textbf{F1 Score (\%)} \\
            \midrule
            0.9 & 91.33 & 90.78 \\
            0.99 & 91.52 & 89.97 \\
            0.999 & \textbf{94.24} & \textbf{93.86} \\
            \bottomrule
        \end{tabular}
    }
    \captionof{table}{$\beta$  - Model Performance Relationship}
    \label{Table_9}
\end{minipage}
\end{table}

In summary, the sensitivity analysis and interleaved experiments reveal the complex coupling relationships among the hyperparameters of the loss functions, resulting in a final accuracy of 94.24\%, which represents a 1.09\% improvement over the 93.16\% achieved in the second stage. Compared with independent optimization of a single loss function, joint optimization of multiple loss functions can more comprehensively reflect the model requirements in practical application scenarios, thereby yielding a more optimal hyperparameter configuration. Therefore, in the actual model tuning process, the interactions among hyperparameters should be fully considered, and multi-parameter joint optimization should be employed to achieve optimal model performance.

\end{appendices}
\end{document}